\documentclass[sigconf,9pt]{acmart}

\usepackage{amsmath,amsfonts}
\usepackage[linesnumbered,ruled,vlined]{algorithm2e}
\usepackage{algorithmic}
\usepackage{graphicx}
\usepackage{textcomp}
\usepackage{xcolor}
\usepackage[caption=false]{subfig}
\usepackage{float}
\usepackage{bm}
\usepackage{soul}
\usepackage{comment}
\usepackage{subfig}
\usepackage{booktabs}
\usepackage{makecell}
\usepackage{colortbl}
\usepackage{xspace}
\usepackage{dblfloatfix}
\usepackage{caption}
\usepackage{marvosym}

\AtBeginDocument{%
  \providecommand\BibTeX{{%
    \normalfont B\kern-0.5em{\scshape i\kern-0.25em b}\kern-0.8em\TeX}}}

\newcommand{\name}{FM-Fi\xspace}

\ifodd 0
\newcommand{\rev}[1]{{\color{blue}#1}} %
\else
\newcommand{\rev}[1]{#1}
\fi

\copyrightyear{2024}
\acmYear{2024}
\setcopyright{rightsretained}
\acmConference[SENSYS '24]{The 22nd ACM Conference on Embedded Networked 
Sensor Systems}{November 4--7, 2024}{Hangzhou, China}
\acmBooktitle{The 22nd ACM Conference on Embedded Networked Sensor 
Systems (SENSYS '24), November 4--7, 2024, Hangzhou, 
China}\acmDOI{10.1145/3666025.3699349}
\acmISBN{979-8-4007-0697-4/24/11}

\begin{document}

\title{Large Model for Small Data: Foundation Model for Cross-Modal RF Human Activity Recognition}

\ifodd 0
\author{Conditionally Accepted Paper \#328 to ACM SenSys 2024\thanks{xxx.}}

\affiliation{
   \institution{{\normalsize
   \textcolor{white}{$^1$ xxx} \\
   \textcolor{white}{$^2$ xxx} \\
   \textcolor{white}{$^3$ xxx} \\
   \textcolor{white}{$^4$ xxx}}
}
    \country{\textcolor{white}{xxx}}
}
\renewcommand{\authors}{Conditionally Accepted Paper \#328 to ACM SenSys}
\renewcommand{\shortauthors}{Conditionally Accepted Paper \#328 to ACM SenSys 2024}
\else
\author{Yuxuan Weng$^1$, \quad Guoquan Wu$^1$, \quad Tianyue Zheng$^{1, 2}$\textsuperscript{\Letter}\thanks{\Letter \quad Corresponding author.},\quad Yanbing Yang$^3$, \quad Jun Luo$^4$}

\affiliation{
   \institution{{\normalsize
   $^1$ Department of Computer Science and Engineering, Southern University of Science and Technology \country{China} \\
   $^2$ Smart City Center, Research Institute of Trustworthy Autonomous Systems, Southern University of Science and Technology  \country{China} \\
   $^3$ College of Computer Science, Sichuan University  \country{China} \\
   $^4$ College of Computing and Data Science, Nanyang Technological University  \country{Singapore}}
}
    \country{Email: \{wengyx, wugq2024, zhengty\}@sustech.edu.cn, yangyanbing@scu.edu.cn, junluo@ntu.edu.sg}
}
\renewcommand{\authors}{Y. Weng, G. Wu, T. Zheng, Y. Yang, and J. Luo}
\renewcommand{\shortauthors}{Y. Weng, G. Wu, T. Zheng, Y. Yang, and J. Luo}
\fi

\begin{abstract}
Radio-Frequency (RF)-based Human Activity Recognition (HAR) rises as a promising solution for applications unamenable to techniques requiring computer visions. However, the \textit{scarcity} of labeled RF data due to their non-interpretable nature poses a significant obstacle. Thanks to the recent breakthrough of \textit{foundation model}s (FMs), extracting deep semantic insights from unlabeled visual data become viable, yet these vision-based FMs fall short when applied to small RF datasets. To bridge this gap, we introduce \name, an innovative cross-modal framework engineered to translate the knowledge of vision-based FMs for enhancing RF-based HAR systems. \name involves a novel cross-modal \textit{contrastive} knowledge distillation mechanism, enabling an RF encoder to inherit the interpretative power of FMs for achieving zero-shot learning. It also employs the intrinsic capabilities of FM and RF to remove extraneous features for better alignment between the two modalities. The framework is further refined through metric-based few-shot learning techniques, aiming to boost the performance for predefined HAR tasks. Comprehensive evaluations evidently indicate that \name rivals the effectiveness of vision-based methodologies, and the evaluation results provide empirical validation of \name's generalizability across various environments.

\end{abstract}



\maketitle

\section{Introduction} \label{sec:intro}

With rapid developments~\cite{hao2018recognizing,dang2020sensor}, Human Activity Recognition (HAR) gains significant interest in smart homes~\cite{chi2018ear,niu2018boosting}, digital healthcare~\cite{wan2020deep,seifert2019toward}, and human-computer interaction~\cite{chen2021magx,rodomagoulakis2016multimodal}. In practice, HAR tasks can be
either contact-based~\cite{bi2019familylog,truong2018capband, ferlini2021eargate} or contact-free~\cite{ding2020rf,chen2021rf}; the latter offers the advantage of not imposing the additional burden and discomfort of wearing devices.
Among all sensing modalities for contact-free HAR, Radio-Frequency (RF) sensing~\cite{palipana2021pantomime,salami2022tesla,hu2023muse,zheng2021more} stands out
by demanding minimal resource for data processing and inference, rendering it ideal for edge device integration. Additionally, it preserves privacy while providing sufficient resolution by capturing only contours without identity-specific features (e.g., facial characteristics and clothing attributes), while being free
of visual constraints~\cite{bansal2020pointillism, boroushaki2021robotic,zheng2021siwa} such as low-light or haze. 
Therefore, RF-HAR is deemed as a promising solution.

Whereas being effective to
specific HAR tasks,  RF sensing is hindered by data scarcity and difficulties in annotation. 
In fact,
comprehensive RF datasets are scarce, and the available ones
often suffer from compatibility issues due to the diversity
in RF devices.
This is caused by
the significant challenges in annotating RF-sensing data~\cite{singh2023depth}:
Unlike 
image data, human annotators find it impossible
to intuitively recognize activities from
RF data, complicating offline annotation. As a result, annotators must resort to online labeling, posing
stringent demands on their skills and increasing the difficulty in
verifying data quality after annotation. Therefore, creating a comprehensive RF-HAR dataset incurs prohibitive
costs yet still lack guaranteed
data reliability, largely confining the adoption of RF sensing in HAR tasks.

The recent advent of Foundation Models (FMs)~\cite{brown2020language,dosovitskiy2020image,ramesh2021zero} presents a promising solution for addressing the scarcity of labeled data in RF-HAR. Due to their large scale and multimodal training on massive datasets, these models have acquired comprehensive
knowledge.
In particular, FMs~\cite{radford2021learning} are trained through an unsupervised process that aligns different data modalities within a high-dimensional space, enabling them to process and understand diverse inputs. Such capabilities enable FMs to generalize across diverse domains, and support applications such as zero-shot image classification~\cite{radford2021learning,wortsman2022robust,esmaeilpour2022zero}, object detection~\cite{minderer2022simple,fang2023eva}, and image generation~\cite{ramesh2021zero,ramesh2022hierarchical}. In particular, the comprehensive knowledge and zero-shot capability of FMs could be crucial to overcome the inherent scarcity of labeled data in RF sensing, and they may also bear the potential to push RF-HAR towards \textit{open-set} recognition~\cite{ren2022delving}. Now the question becomes: \textit{can FMs be harnessed to interpret RF-HAR data?}
A valid answer to this question is essential for advancing RF-HAR towards practical adoption.

\begin{figure}[t]
    \centering 
    \setlength{\abovecaptionskip}{8pt}  
    \includegraphics[width=0.98\linewidth]{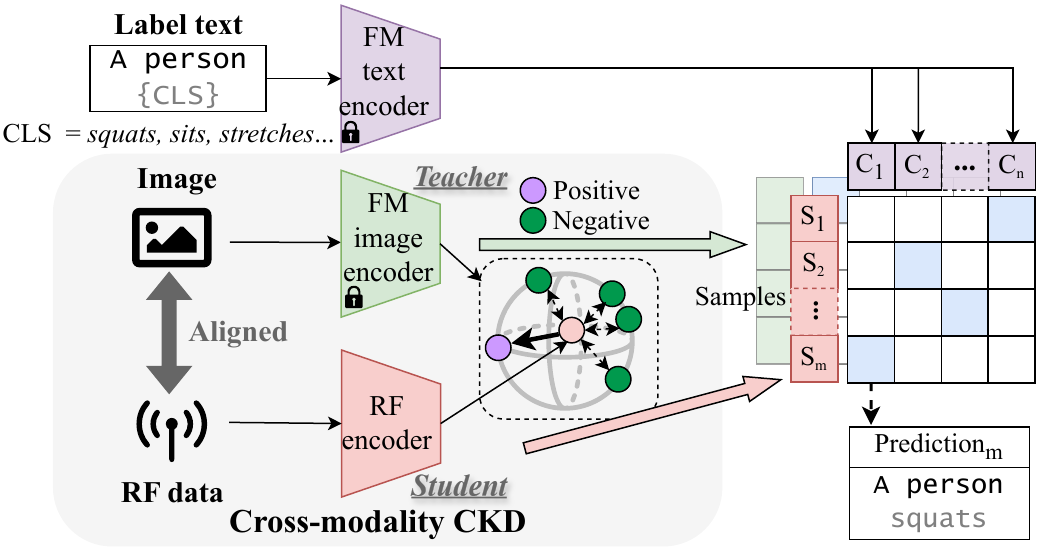}
    \caption{Overview of \name.} 
    \label{fig:teaser} 
    \vspace{-1em}
\end{figure}

Despite the potential of FMs in various domains, applying them to interpret RF-HAR data presents several unique challenges. First, the majority of existing FMs have been primarily developed for tasks in computer vision (CV)~\cite{dosovitskiy2020image} and natural language processing (NLP)~\cite{brown2020language,radford2021learning}, thus limiting their direct applicability to RF-HAR. Although cross-modal knowledge distillation (KD)~\cite{hinton2015distilling} paves the way for knowledge transfer from image to RF modality, their efficacy in adapting to the structured embeddings of FMs remains unexplored.
Second, the image and RF modalities exhibit inherent feature discrepancies. Specifically, the image modality include extraneous background details that obscure HAR-relevant information, whereas the RF modality often features irrevelant static backgrounds. This misalignment significantly challenges effective modality integration. Third, while FMs produce informative embeddings, their optimal use in HAR requires further fine-tuning. However, this fine-tuning process is hindered by the scarcity (or void) of labeled data.%

To tackle these challenges, we design \name, a cross-modal framework that distills the knowledge from FMs to the RF modality, as illustrated in Figure~\ref{fig:teaser}. First, given that conventional KD does not consider the structures and interdependencies among the embeddings generated by FMs, we design a novel \textit{contrastive knowledge distillation} (CKD) for transferring knowledge from FM to \rev{the neural model for the RF modality. 
As opposed to conventional KDs, our CKD stems from the mutual information between the embeddings of two modalities: since the interdependency among the embeddings' elements is captured as a form of ``information'', they can thus be better preserved during distillation.}
Second, \name employs the intrinsic capabilities of FM and RF to remove extraneous background features, thus enabling better alignment between the two modalities. \rev{In particular, the semantic space of FM is leveraged to score vision features, and the physical properties of the RF modality are explored to filter static and dynamic backgrounds.}
Finally, \name harnesses a minimal set of annotated data to fine-tune its model via metric-based few-shot learning, enhancing already achieved zero-shot classification to fit specific HAR tasks. The synergy of these three mechanisms sets the stage for the RF encoder to acquire the full capabilities of the FMs, while opening the way for approaching open-set HAR given the constant improvement of FMs. In summary, our key contributions are: 
\begin{itemize}
    \item To the best of our knowledge, \name\ is the first cross-modal distillation system specifically designed from vision FMs to RF model for zero/few-shot HAR tasks.
    \item We develop a CKD mechanism to accommodate FM's intrinsic embedding dependencies, enabling successful knowledge transfer from FMs to RF modality.
    \item We design extraneous feature elimination methods tailored to image and RF modalities, achieving a feature-aligned vision-RF dataset.
    \item We design a metric-based few-shot learning mechanism to fine-tune the RF encoder, thereby adapting and enhancing it for specific closed-set HAR tasks.
    \item We construct an \name\ prototype and evaluate it with extensive experiments: the promising results confirm that \name enables high-performance RF-HAR for both zero-shot and few-shot HAR tasks.
\end{itemize}

In the following, \S~\ref{sec:background_motivation} introduces the background and motivation of \name. \S~\ref{sec:system_design} presents the system design of \name. \S~\ref{sec:evaluation} introduces the datasets, system implementation, and experiment setup, before reporting the evaluation results. Related and future works are discussed in \S~\ref{sec:discussion}. Finally, \S~\ref{sec: conclusion} concludes the paper with future directions.
\section{Background and Motivations} \label{sec:background_motivation}
In this section, we introduce the background of FM for HAR and the motivations of \name's design.

\subsection{FM for HAR}
FMs represent a novel category of large-scale neural networks trained on datasets comprising billions of samples. The training occurs across multiple GPUs over a span of several weeks. Their rapid adoption across various domains, such as CV (e.g., DALLE~\cite{ramesh2021zero} for image generation), NLP (e.g., GPT~\cite{brown2020language} for chatbot), and multimodal applications (e.g., CLIP~\cite{radford2021learning} for image semantics understanding), have demonstrated their extensive capabilities. The enhanced image understanding in FMs is facilitated by the adoption of transformer~\cite{{vaswani2017attention, dosovitskiy2020image}} architecture as encoders, which enable the derivation of complex representations. Additionally, contrastive learning~\cite{chen2020simple,he2020momentum} has been exploited to align embeddings across different modalities, integrating visual data with semantic insights. Last but not least, the training methodology benefits from the use of unlabeled image-text pairs, allowing for the creation of large-scale training datasets. All these properties have enabled FMs to accurately align image and label embeddings for classification tasks regardless of sample dependency. 
\begin{figure}[b]
    \setlength\abovecaptionskip{8pt}
    \vspace{-1em}
	   \captionsetup[subfigure]{justification=centering}
		\centering
		\subfloat[On image modality.]{
		  \begin{minipage}[b]{0.48\linewidth}
		        \centering
			    \includegraphics[width = 0.98\textwidth]{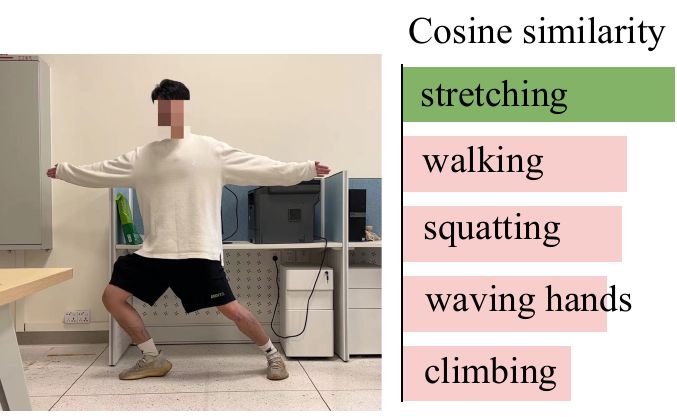}
                    
			\end{minipage}
			\label{subfig:fm_for_har_a}
		}
		\subfloat[On RF modality.]{
		    \begin{minipage}[b]{0.48\linewidth}
		        \centering
			    \includegraphics[width = 0.98\textwidth]{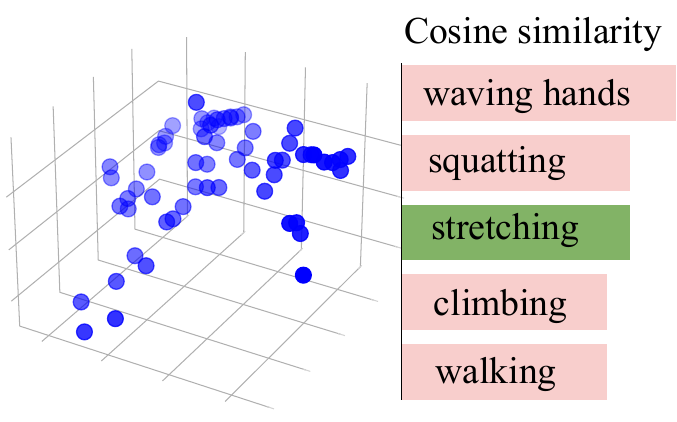}
                    
			\end{minipage}
			\label{subfig:fm_for_har_b}
		}
		\caption{Performance of FM for HAR.}
		\label{fig:FM_for_HAR}
		\vspace{-.5ex}
\end{figure}

The interpretive power of FMs makes them ideal tools for conducting HAR. For instance, when analyzing an image of a person stretching, as depicted in Figure~\ref{subfig:fm_for_har_a}, the CLIP model can accurately assess the similarity between the embedding of the image and the text, thereby achieving zero-shot HAR. However, the application of FMs, initially trained on vision-text data, to RF data introduces considerable challenges. This difficulty arises from the inherent abstractness of RF signals. As illustrated in Figure~\ref{subfig:fm_for_har_b}, directly applying the CLIP model has falsely identified the activity \textit{stretching} captured by a mmWave (in the form of point cloud akin to image pixels) as \textit{waving hands}. This limitation underscores the necessity of novel methods for RF data processing to extend the applicability of FMs beyond visual data.

\subsection{Why Conventional KD Fails for FMs?} \label{subsec:kd_fail}
\begin{figure}[h]
    \setlength\abovecaptionskip{8pt}
    \vspace{-1.2em}
	   \captionsetup[subfigure]{justification=centering}
		\centering
		\subfloat[Zero-shot HAR comparison.]{
		  \begin{minipage}[b]{0.48\linewidth}
		        \centering
			    \includegraphics[width = 0.96\textwidth]{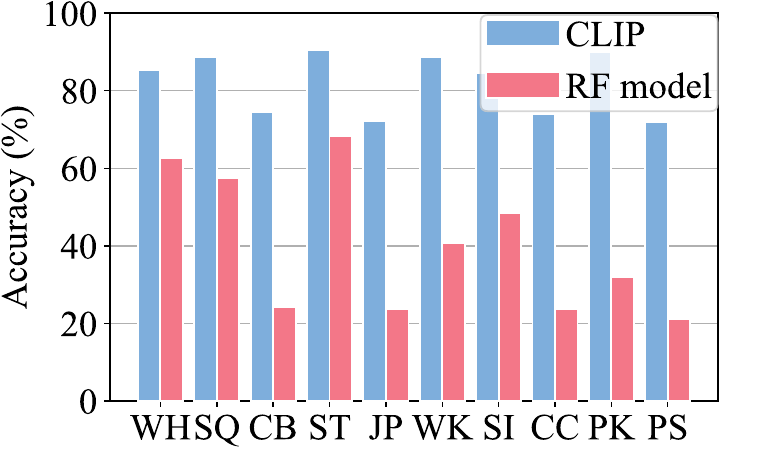}
                    
			\end{minipage}
			\label{subfig:kd_acc}
		}
		\subfloat[$|\mathbf{R}^\mathrm{IM}-\mathbf{R}^\mathrm{RF}|$.]{
		    \begin{minipage}[b]{0.48\linewidth}
		        \centering
			    \includegraphics[width = 0.96\textwidth]{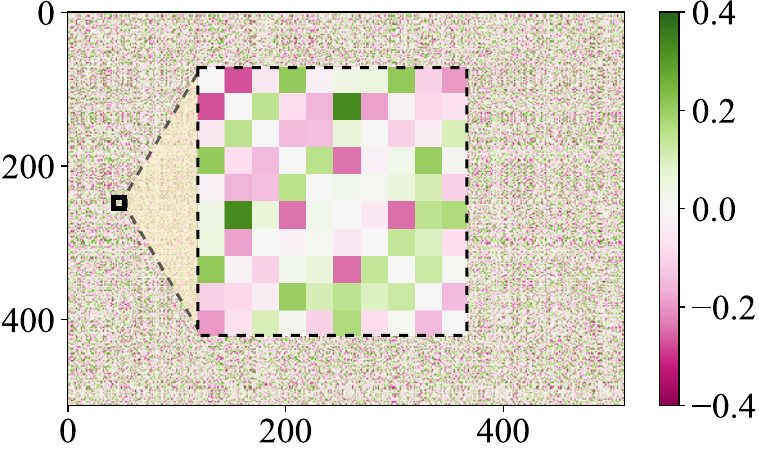}
                    
			\end{minipage}
			\label{subfig:kd_map}
		}
		\caption{Conventional KD performance.}
		\label{fig:kd_fail}
    \vspace{-.8em}	
\end{figure}

One viable approach for utilizing FMs for HAR is KD; it involves transferring the knowledge from an FM to RF model by aligning their output embeddings, where we utilize the mean squared error (MSE) loss for an element-wise comparison of embeddings between image and RF modalities. We employ a synchronized image-RF dataset in our experiment, \rev{whose classes will be detailed in \S~\ref{subsec:experiment_setup},} to assess
the zero-shot HAR performance, by comparing a CLIP model with an RF model trained via a standard KD~\cite{hinton2015distilling}. One may readily observe that a naive application of KD on FMs leads to inferior performance, as depicted in Figure~\ref{subfig:kd_acc}. Specifically, for 10-class classification, the CLIP-trained RF encoder achieves an average accuracy slightly above 40\%, whereas that achieved by the baseline CLIP exceeds 80\%.

To understand KD's ineffectiveness, we explore the interdependencies among elements of the output embeddings. We compute the correlation matrices, $\mathbf{R}^\mathrm{IM}$ for the FM (processing the image modality) and $\mathbf{R}^\mathrm{RF}$ for the RF model, respectively. By subtracting $\mathbf{R}^\mathrm{RF}$ from $\mathbf{R}^\mathrm{IM}$, we obtain a difference matrix as shown in Figure~\ref{subfig:kd_map}. One may readily observe that the correlation difference of the two embeddings can be significant and reach up to 0.4. This finding reveals the limitation of KD: while it aligns the embeddings from the FM and RF model on an element-wise basis, it fails to account for the interdependencies among the elements of the FM's embeddings~\cite{peng2019correlation}. The interdependency is especially important for HAR, it is essential that latent factors representing the human subject, various body parts, and activity states should be related and active, while other irrelevant factors should also be related but suppressed. We forward reference to Figure~\ref{subfig:cm_ckd_b} in \S~\ref{subsec:Cross-Modality Distillation} for a better correlation matrix difference that better captures the interdependencies among the elements in the embeddings.

\subsection{Effect of Extraneous Feature} \label{subsec:Effect of Extraneous Feature}
\begin{figure}[t]
\centering 
\includegraphics[width=0.95\linewidth]{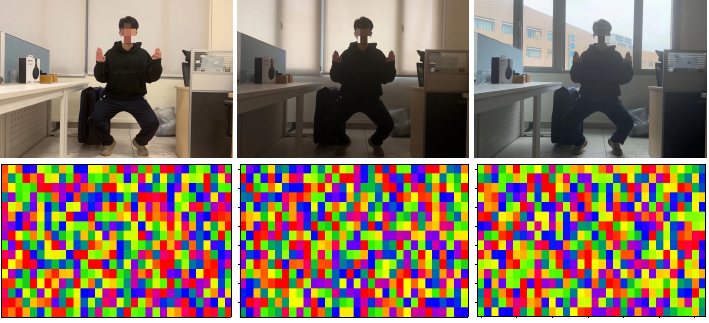}
\caption{Minor background variations significantly alter the output embeddings of FM.} 
\label{fig:fig5} 
\vspace{-1em}
\end{figure}

To successfully transfer knowledge from the image to RF modality, alignment between the two modalities is crucial. Both modalities, however, contain extraneous features; for instance, images may include irrelevant lighting and background objects, while RF data may be influenced by static backgrounds. As demonstrated in Figure~\ref{fig:fig5}, minor variations in background features, such as lighting and curtains (as illustrated in the upper row), significantly affect the embeddings (in the lower row), leading to instability. This instability is presumed to affect the RF modality as well. Furthermore, there is no straightforward one-to-one correspondence between the embeddings of image and RF modalities due to their not sharing an identical set of features. Consequently, these extraneous features hinder the knowledge transfer from image to RF modality, necessitating the development of a method to efficiently eliminate such features.

\section{System Design} \label{sec:system_design}

\begin{figure*}[t]
    \centering
    \includegraphics[width=0.97\linewidth]{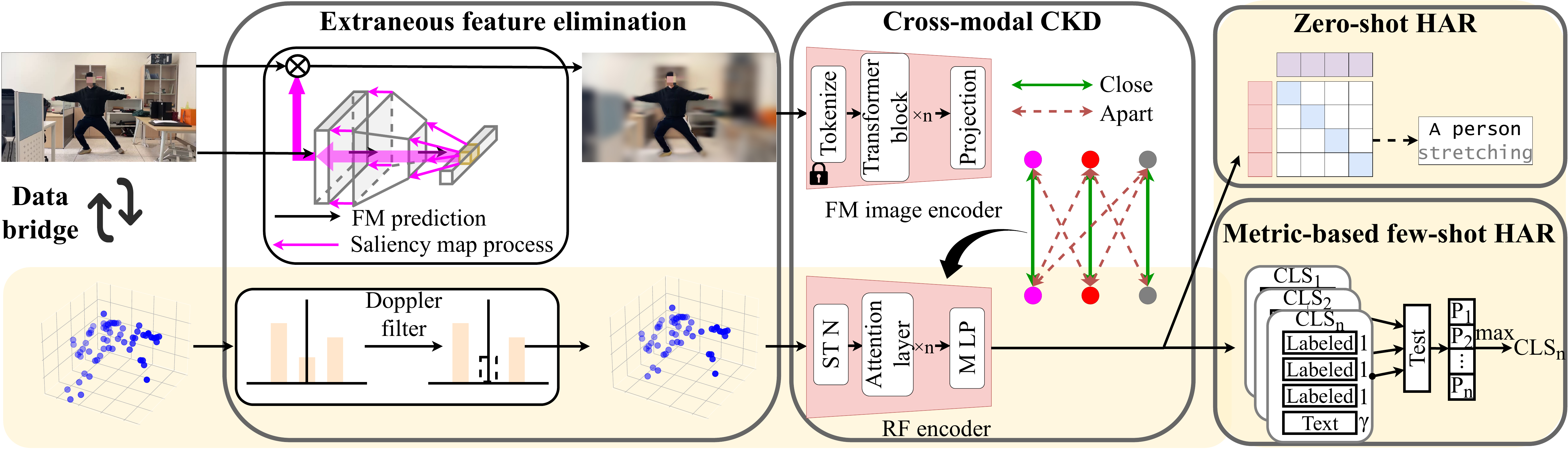}
    \caption{Overall design of \name.}
    \label{fig:fig6}
    \vspace{-.5em}
\end{figure*}

Based on the discussions in \S~\!\ref{sec:background_motivation}, we hereby present \name with five innovative components: i) an RF encoder that helps encoding information from the RF point clouds, ii) a cross-modal CKD framework for transferring semantic representations from visual feature maps to RF-based models, iii) a multimodal data alignment module that eliminates extraneous features, thereby improving HAR knowledge integration across modalities, iv) a zero-shot HAR mechanism relying on learned associations between the semantics of both RF and (FM's) text modalities, and v) a metric-based few-shot learning network enabling \name to quickly adapt to various closed-set HAR tasks with few labeled examples. In the following, we elaborate on each component, given the overall design depicted in Figure~\ref{fig:fig6}.

\subsection{RF Encoder} \label{ssec:rf_encoder}
\begin{figure}[b]
\centering 
\vspace{-2ex}
\includegraphics[width=\linewidth]{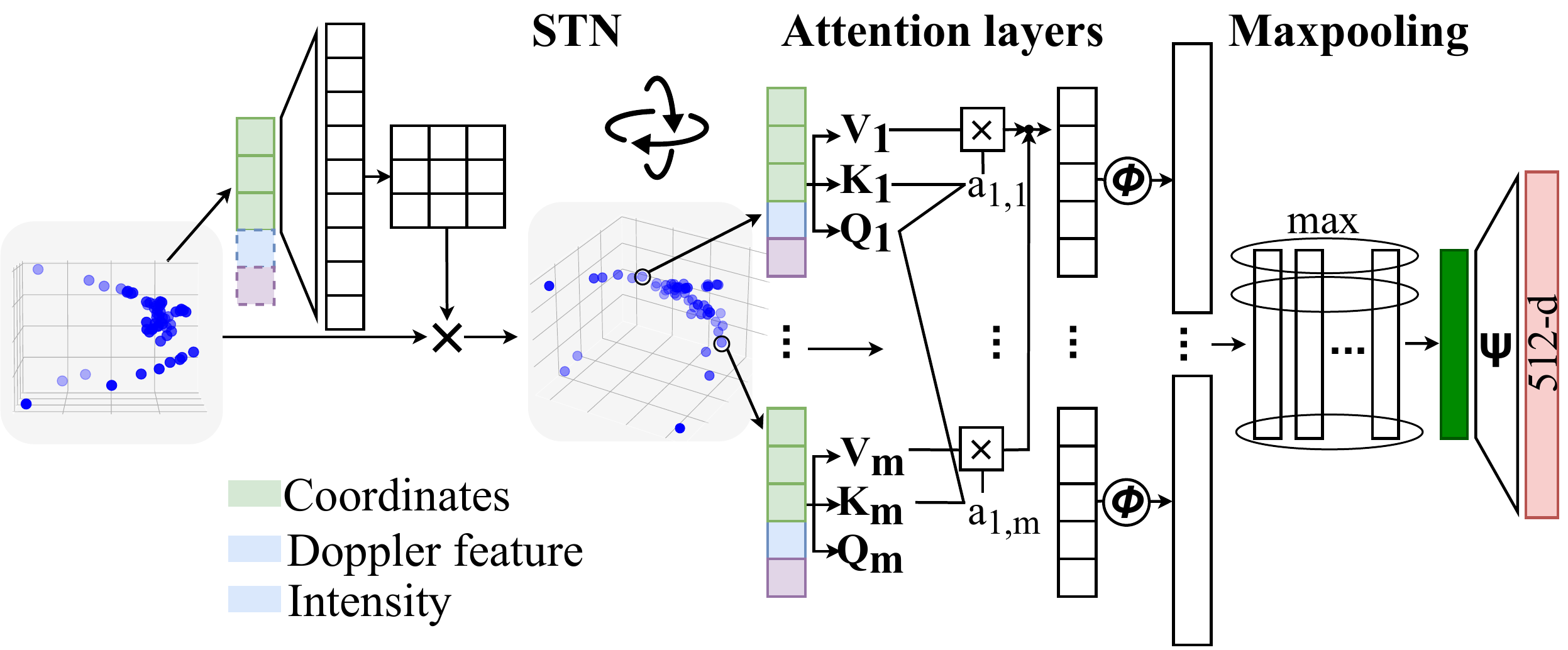}
\caption{RF encoder for cross-modal distillation.} 
\label{fig:fig7} 
\vspace{-.5ex}
\end{figure}

The mmWave data collected for this study is presented as a point cloud, containing information of coordinates, Doppler frequency, and intensity, \rev{each of which is indispensable for HAR analysis. Specifically, the point cloud coordinates provide valuable insights into human posture, while intensity reveals the reflection characteristics, and Doppler frequency offers critical dynamic information regarding motion. Before being processed by the neural network, the point cloud undergoes preprocessing, during which their centroid is translated to the origin, effectively eliminating any translational biases. Drawing upon these rich features, we develop a robust RF encoder for extracting meaningful RF embeddings.}
Contrary to the inherent order of image pixels, point cloud data is characterized by an absence of order. Furthermore, the coordinates of a point cloud depend on the selected coordinate system. However, neither changing the point order nor the coordinate system should affect the feature extraction outcome. To address these challenges, we revamp the design of PointNet~\cite{qi2017pointnet} to accommodate the properties of mmWave data, as shown in Figure~\ref{fig:fig7}. \name's RF encoder includes a spatial transformation network (STN) $\mathcal{T}$, attention layers, and a maxpooling module. STN aims to learn a $3 \times 3$ rotation-scaling matrix $\mathbf{W}_T$, implementing a transformation on each point as $\mathbf{x}' = \mathbf{W}_T \cdot \mathbf{x}$, where $\mathbf{x}$ and $\mathbf{x}'$ represent the original and transformed coordinates, respectively. To derive $\mathbf{W}_T$, the point cloud undergoes processing through convolutional layers and fully connected layers, outputting a 9-dimensional vector reshaped into a $3 \times 3$ matrix. Through this process, the STN captures the relationship between the point cloud's global distribution and implicit viewpoint information, as $\mathbf{W}_T = \mathcal{T}(\mathbf{x})$. This transformation standardizes the point cloud, and improves its robustness against geometric variations.

It is important to note that, in addition to spatial coordinates $(x, y, z)$, mmWave point clouds incorporate two additional features: Doppler frequency and intensity. The Doppler feature provides information about the moving velocity of targets, while intensity is indicative of their distance and material properties. These two features are essential for HAR and are consequently concatenated with the three-dimensional coordinates after STN processing. The resulting feature vector, now enriched with the transformed coordinates and the two additional features, is fed into a module $\mathcal{A}$, consisting of several self-attention layers. This module is tasked with selectively weighting individual points \rev{within the global context, thereby} effectively filtering out those irrelevant to HAR. The design details of this module are further discussed in \S~\ref{subsubsec:RF Modality}.

We then pass the enriched feature vectors through a multilayer perceptron (MLP) $\phi$ for dimensionality expansion, after which the updated coordinates are processed by a maxpooling module. This module selects the maximal value across all points for each element of the embedding, a process that remains invariant to the order of point inputs and equally emphasizes every point in the space. It should be noted that this step processes the point cloud as a whole, rather than focusing on individual points. Subsequent to another MLP, denoted as $\psi$, the output of the RF encoder is mapped to a 512-dimensional vector. In summary, the process of point cloud processing can be expressed as follows:
\begin{align}
    \mathbf{E}^{\mathrm{RF}} = \psi\left(\textstyle{\underset{i=1 \ldots N}{\mathrm{maxpooling}}} \left( \phi\left( \mathcal{A}\left(\mathcal{T}(\mathbf{X}_\mathrm{RF_i})\cdot \mathbf{X}_\mathrm{RF_i} \right) \right) \right) \right).
\end{align}
The 512-dimensional output of the encoder guarantees compatibility with the output from the FM image encoder.

\subsection{Cross-Modal CKD} \label{subsec:Cross-Modality Distillation}

\rev{Synchronized vision and RF modalities capturing the same scene offer closely related physical information, such as spatial structure, contours, and dynamic information. As a result, the gap between their semantic embedding spaces can be potentially bridged using knowledge distillation~\cite{fan2020home}. }
The first step in conducting KD from FM to RF models involves constructing a data bridge to link the image and RF modalities. Given the scarcity of annotated data, highlighted in Section~\ref{subsec:kd_fail}, this bridge only employs unlabeled synchronized data gathered from a pair of camera and radar sensor. Specifically, it comprises two data types: i) unstructured data from everyday spontaneous activities, and ii) rehabilitation activity data. The former provides a large amount of data that captures real-world complexities, aiding in model generalization; while the latter includes a wide range of body movements encompassing rare movement cases, thereby offering extensive body variation and motion diversity. This comprehensive data bridge selection ensures the subsequent KD process transcends mere recognition of specific movements and body parts under few environments.

Specifically, we collect datasets consisting of paired image and RF data, represented as ${(\mathbf{X}^\mathrm{IM}_i, \mathbf{X}^\mathrm{RF}_i)}$, where $i = 1, \cdots, N$. These datasets are gathered from the same scenes to bridge the modalities. For each modality, data is processed by the corresponding encoder, producing embeddings $\mathbf{E}^\mathrm{IM}$ and $\mathbf{E}^\mathrm{RF}$.
\rev{While the representations of different modalities share some common information, they do have some differences that cannot be aligned. This means relying solely on rigid metrics like the Euclidean distance in traditional KD is insufficient, as discussed in \S~\ref{subsec:kd_fail}. Instead, we employ the mutual information between modalities as the starting point for deriving contrastive knowledge distillation (CKD) method.}
This method is better at handling interdependencies within the embedding elements, \rev{which are crucial for storing information of the embeddings.} Specifically, to distill the interdependency information critical for HAR, CKD maximizes the lower bound of the mutual information $MI$ between the image and RF embeddings $\mathbf{E}^{\mathrm{IM}}$ and $\mathbf{E}^{\mathrm{RF}}$. The mutual information is defined as: $MI(\mathbf{E}^{\mathrm{IM}}; \mathbf{E}^\mathrm{RF}) = \mathbb{E}_{p(\mathbf{E}^{\mathrm{IM}}, \mathbf{E}^\mathrm{RF})} \left[ \log \frac{p(\mathbf{E}^\mathrm{RF}|\mathbf{E}^{\mathrm{IM}})}{p(\mathbf{E}^\mathrm{RF})} \right]$. 
Assuming $ \mathbf{E}^\mathrm{RF} $ follows a uniform distribution (i.e., $ p(\mathbf{E}^\mathrm{RF}) = \frac{1}{N} $), we have:
\begin{align}
    MI(\mathbf{E}^{\mathrm{IM}};\mathbf{E}^\mathrm{RF}) = \mathbb{E}_{p(\mathbf{E}^{\mathrm{IM}}, \mathbf{E}^\mathrm{RF})} \left[ \log p(\mathbf{E}^\mathrm{RF}|\mathbf{E}^{\mathrm{IM}}) \right] + \log N. \nonumber
\end{align}
The conditional probability $ p(\mathbf{E}^\mathrm{RF}|\mathbf{E}^{\mathrm{IM}}) $ is estimated as:
\begin{align}
    p(\mathbf{E}^\mathrm{RF}|\mathbf{E}^{\mathrm{IM}}) \geq \frac{\exp(\mathrm{sim}(\mathbf{E}^{\mathrm{IM}}, \mathbf{E}^\mathrm{RF}))}{\sum_{\mathbf{E}^\mathrm{RF'} \in \mathcal{P}} \exp(\mathrm{sim}(\mathbf{E}^{\mathrm{IM}}, \mathbf{E}^\mathrm{RF'}))}. \nonumber
\end{align}%
where $ \mathrm{sim}(\cdot) $ measures the similarity  between $\mathbf{E}^\mathrm{IM}$ and $\mathbf{E}^\mathrm{RF}$, and $\mathcal{P}$ is the set of all possible samples $\mathbf{E}^\mathrm{RF'}$. Therefore we have $MI(\mathbf{E}^{\mathrm{IM}};\mathbf{E}^\mathrm{RF}) \geq \log N - \mathcal{L}_\mathrm{CKD}$, where
\begin{align}
    \mathcal{L}_\mathrm{CKD} = -\mathbb{E}_{p(\mathbf{E}^\mathrm{IM}, \mathbf{E}^\mathrm{RF})}\left[\log \frac{\exp(\mathrm{sim}(\mathbf{E}^\mathrm{IM}, \mathbf{E}^\mathrm{RF}))}{\sum_{\mathbf{E}^\mathrm{RF'} \in \mathcal{P}} \exp(\mathrm{sim}(\mathbf{E}^\mathrm{IM}, \mathbf{E}^\mathrm{RF'}))}\right], \nonumber
\end{align}
where $ \mathrm{sim}(\cdot) $ is defined as $\langle \cdot, \cdot \rangle/\tau$, with $\langle \cdot, \cdot \rangle$ being the cosine similarity, and $\tau$ being the temperature scaling parameter.
\rev{It should be noted that, while the mathematical structure of CKD loss may resemble conventional contrastive losses, its underlying computation process is considerably different. First, the positive samples in CKD are drawn from the teacher modality's embeddings, which eliminates the need for data augmentation. Second, the student modality interacts solely with the teacher modality for comparison, bypassing intra-modal comparisons and significantly reducing computational overhead. Lastly, CKD leverages cosine similarity for measuring similarities of the embeddings, thereby eliminating the reliance on a critic model, as required by another cross-modal distillation baseline CRD~\cite{tian2019contrastive}.}

\begin{figure}[b]
    \setlength\abovecaptionskip{8pt}
    \vspace{-1em}
	   \captionsetup[subfigure]{justification=centering}
		\centering
		\subfloat[Basic idea.]{
		  \begin{minipage}[b]{0.495\linewidth}
		        \centering
			    \includegraphics[width = 0.96\textwidth]{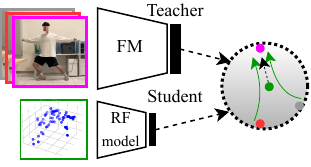}
			\end{minipage}
			\label{subfig:cm_ckd_a}
		}
		\subfloat[$|\mathbf{R}^\mathrm{IM}-\mathbf{R}^\mathrm{RF}|$.]{
		    \begin{minipage}[b]{0.495\linewidth}
		        \centering
			    \includegraphics[width = 0.96\textwidth]{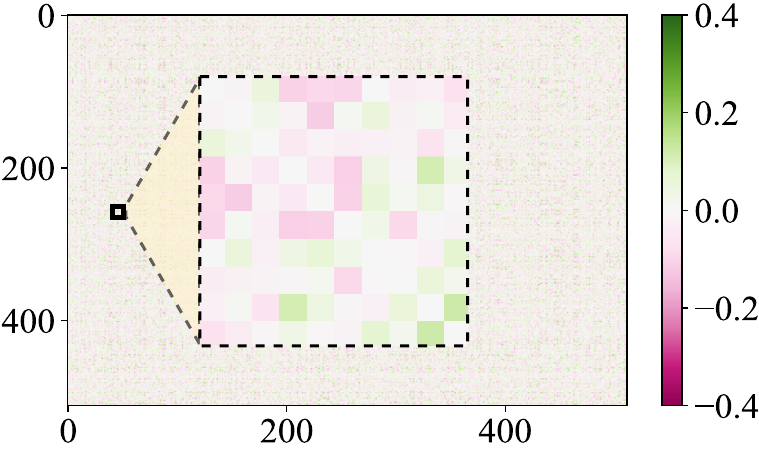}
                    
			\end{minipage}
			\label{subfig:cm_ckd_b}
		}
		\caption{Cross-modal CKD.}
		\label{fig:cm_ckd}
    \vspace{-.5ex}	
\end{figure}

As shown in Figure~\ref{subfig:cm_ckd_a}, CKD reduces the distance between embeddings of positive RF-image pairs, while simultaneously increasing the separation between negative pairs within the embedding space. This contrastive method enhances the distillation process by more effectively capturing the interdependencies among the embedding elements. Additionally, Figure~\ref{subfig:cm_ckd_b} shows a significant reduction of 0.2 on average, in the difference between the correlation matrices of the FM and RF, denoted as $|\mathbf{R}^\mathrm{IM}-\mathbf{R}^\mathrm{RF}|$, when utilizing CKD. This contrasts with the outcomes observed with traditional KD, as depicted in Figure~\ref{subfig:kd_map}. This observation underscores CKD's superiority in aligning the structural characteristics of the embeddings across diversified modalities.

\subsection{Extraneous Feature Elimination} \label{subsec:Extraneous Feature Elimination}
As described in \S ~\ref{subsec:Effect of Extraneous Feature}, our goal is to remove extraneous features to allow CKD to focus on HAR-relevant features. To improve interpretability, facilitate better integration, and reduce the consumption of computational resources, we perform feature elimination by utilizing signal properties and FM's knowledge without any extra models. %

\subsubsection{Image Modality} \label{subsubsec:feature elimination image}

\begin{figure}[t]
\centering 
\includegraphics[width=0.98\linewidth]{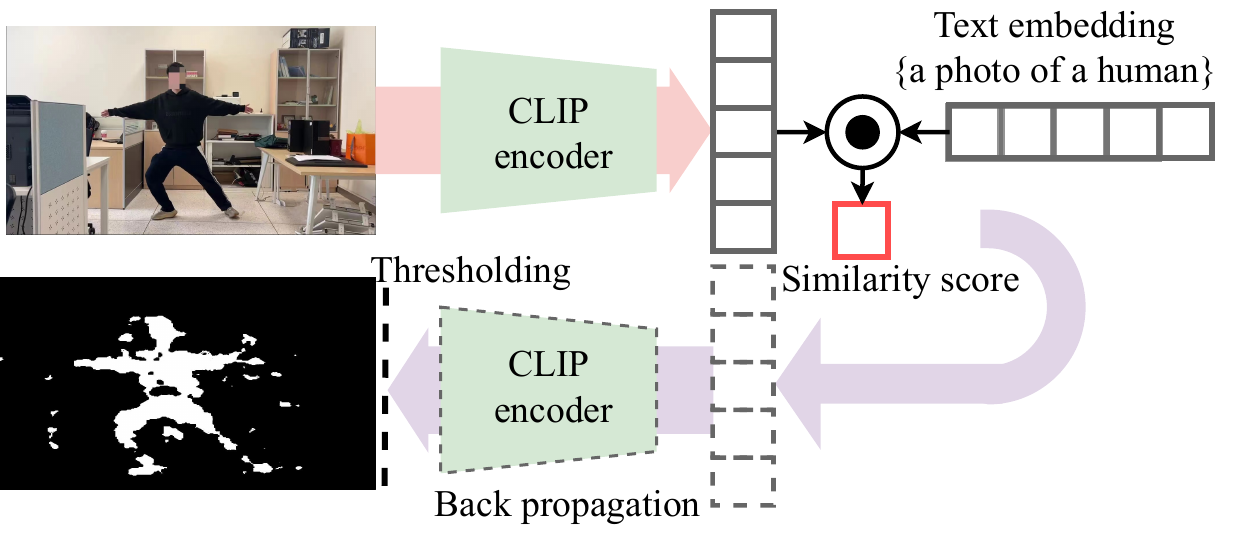}
\caption{Generation of (threshold) HAR saliency map.} 
\label{fig:fig8} 
\end{figure}

Instead of employing an extra segmentation model for annotation, we employ the image and text encoders from the teacher model in the CKD framework to generate saliency maps~\cite{simonyan2014deep}. 
A saliency map has the same dimensions as the input image, where each element's magnitude quantifies the importance of the corresponding pixel in determining the model's predictive output of a human. It enables the isolation of image regions that are pertinent to human activity, allowing for the exclusion of non-essential features. 
\rev{Compared with other segmentation approaches, \name eliminates the need for additional neural networks, and avoids potential issues that could arise from incompatible weighting method of input features by non-CLIP neural networks.}
As shown in Figure~\ref{fig:fig8}, an image processed through the CLIP encoder produces an embedding vector that encapsulates the spatial and contour information of the human body. We compute the similarity score $\mathcal{S}$ by comparing this vector with the text embedding of \textit{``a photo of a human''}, which provides a structural interpretation of these attributes. Following this, we determine the gradient of $\mathcal{S}$ with respect to each input feature of the original image. The aggregate of gradients within the designated target region $T$ signifies the relevance of that feature to the model's output. The saliency map $M$'s individual elements can be obtained as follows:
\begin{align} \label{eq:saliency}
    M(u, v)=\textstyle{\sum_{(u_t, v_t) \in T}}~~\partial \mathcal{S}/\partial I(u_t, v_t),
\end{align}
where $(u, v)$ represents the pixel coordinates and $I$ the original image. In practice, backpropagation can be applied to the scores, generating a chain of gradients across layers equivalent to the gradient in Eqn.~\eqref{eq:saliency}. This process infers the critical elements within each layer that the model deems essential for discrimination, culminating in the identification of salient features within the input. As a result, the processed saliency region can be expressed as $\mathcal{F}_\mathrm{sal}(u, v)=\mathbb{I}\left[M'(u, v)>\lambda\right],$ where $\mathbb{I}(\cdot)$ is the indicator function and $\lambda$ is the threshold, and $M'$ denotes the normalized saliency map. Elements exceeding the threshold retain their original pixel values, whereas those below the threshold are subjected to Gaussian kernel blurring. 
The extensive knowledge and complex architecture of the FM contributes to its accurate outputs and reliable reasoning process. As a result, saliency maps obtained from it efficiently concentrate on the relevant features in images.

\subsubsection{RF Modality} \label{subsubsec:RF Modality}
Within the RF modality, we first eliminate static backgrounds based on the intrinsic physical properties of the data through a deductive approach. 
Taking mmWave radar as an example, the sensor emits electromagnetic waves in the range of 30-300~\!GHz and receives the waves reflected by objects. The raw baseband data collected can be processed to derive information such as distance, angle, and velocity, which can be further transformed into machine learning-friendly input features, such as point clouds. Specifically, distance is calculated based on the time interval between the emission and reception of the waves, while the angle of an object can be estimated using multiple receiving antennas. 
The velocity of an object is inferred through the Doppler effect, which dictates that the frequency shift of the radar waves can be formulated as $f_d = \frac{2v}{c}f_0$, where $f_d$ is the frequency difference between the reflected and emitted waves, $f_0$ is the frequency of the transmitted signal, $c$ is the speed of light, and $v$ is the velocity of the target object relative to the radar sensor. Signals in the point cloud with $f_d=0$, indicative of static backgrounds, are filtered out to isolate dynamic subjects.
\rev{It should be noted that, while it is theoretically possible to mistakenly filter out purely tangential activities (characterized by a Doppler velocity of zero), the likelihood of such occurrences is minimal due to the diversity of MIMO sensors and the abundance of data points associated with a single human subject in real-world scenarios.}

However, in addition to the static background, the scene may also contain other objects that are irrelevant to HAR (e.g., a moving pet).
These objects can be identified by integrating the aforementioned Doppler data, with intensity data that conveys material characteristics, in conjunction with the coordinates derived from point clouds. As detailed in \S ~\ref{ssec:rf_encoder} \rev{and Figure~\ref{fig:fig7}}, the RF encoder firstly transforms the three-dimensional coordinates through rotation and scaling, which are then concatenated with Doppler and intensity to form an enriched point cloud feature vector. To eliminate HAR-irrelevant objects, we introduce a self-attention-based module within the RF encoder which allows the RF model to autonomously discern points of interest \rev{within a global context} by learning from background-free FMs during the cross-modal learning process. 
More specifically, within each layer, we optimize three shared-weight matrices $\mathbf{W}_q$, $\mathbf{W}_k$, and $\mathbf{W}_v$ across all points. The 5-dimensional feature vector $\mathbf{p}$ of each point is transformed into corresponding query $\mathbf{Q} = \mathbf{W}_q \cdot \mathbf{p}$, key $\mathbf{K} = \mathbf{W}_k \cdot \mathbf{p}$, and value $\mathbf{V} = \mathbf{W}_v \cdot \mathbf{p}$. The weighted point vector $\mathbf{p}'$ can be calculated as $\mathbf{p}' = \text{Attention}(\mathbf{Q}, \mathbf{K}, \mathbf{V}) = \text{softmax}\left(\frac{\mathbf{Q}\mathbf{K}^T}{\sqrt{d_k}}\right)\mathbf{V}$.
The self-attention mechanism enables the model to learn to score based on $\mathbf{W}_q$, $\mathbf{W}_k$, and $\mathbf{W}_v$, which prioritizes points relevant to HAR by considering inter-point relationships and the influence of individual points on the global outcome. Our self-attention module explicitly expresses the focus on specific regions, thereby offering enhanced interpretability. Moreover, in contrast to standard MLPs with fixed architectures, it dynamically adjusts the weighting of points according to the input data distribution, thereby increasing the reliability of the model's decision-making.

\subsection{Zero-Shot HAR}
Given that FMs are not trained by simply mapping samples to fixed categories, but rather by understanding the relationship between image content and arbitrary textual descriptions, they are adept at handling certain zero-shot tasks, capable of accurately identifying categories not present in the training set. For instance, CLIP leverages image and descriptive text matching to categorize 1,000 classes in ImageNet within a zero-shot manner. RF models trained under its supervision exhibit similar classification capabilities. Specifically, for any HAR class described in natural language, we can embed it into an appropriate prompt, such as ``A person \{$CLS$\}'', where $CLS$ denotes action like ``walking'' or ``squatting''. 
Subsequently, the text description of this class is divided into individual words, known as tokens. Each token is then transformed into a corresponding numerical value that aligns with a vocabulary defined during the encoder's training phase. As a result, the CLIP text encoder processes these numerical representations rather than the original natural language to generate a 512-dimensional text embedding. 

Following cross-modal CKD, the RF encoder has been endowed with the capability of the vision FMs to embed spatial information into the semantic space. Consequently, it can embed RF data into 512-dimensional vectors, congruent with the previously described text embedding structure. The cosine similarity between embedding vectors from different modalities serves as the criterion for their congruence, with the highest scoring category being selected for prediction $\hat{I}$. The prediction process can be formulated as $\hat{I}  = \arg\underset{\mathbf{E}^\mathrm{TX}}{\max} \left( \frac{\mathbf{E}^\mathrm{TX} \cdot \mathbf{E}^\mathrm{IM}}{\| \mathbf{E}^\mathrm{TX} \| \| \mathbf{E}^\mathrm{IM} \|} \right),$ \rev{where $\mathbf{E}^\mathrm{TX}$  represents the text embedding of the label.} 
To optimize computation, we stack the text embeddings of all candidate labels to create a matrix $W_{\text{zero-shot}} \in \mathbb{R}^{512 \times k}$, whereby
$\text{score} = W_{\text{zero-shot}} \cdot \mathbf{E}^\mathrm{IM}$. 
Given that each text embedding is normalized, we identify the category corresponding to the highest score to make prediction. This matrix computation methodology prevents redundant calculations and enhances the overall efficiency.

\subsection{Metric-Based Few-Shot HAR}

While zero-shot learning adequately addresses most HAR tasks, for especially challenging ones characterized by less distinct language descriptions, we introduce an additional few-shot learning module. This module adopts a metric-based approach utilizing a non-parametric method to predict labels in the query set based on a weighted sum of true labels in the support set. In contrast to conventional metric-based learning, \name's embedding space is semantically rich. As such, we enhance the performance of classification by utilizing the label text embeddings generated by FMs, further exploiting the semantic information they contain. Specifically, we employ cosine similarity as our metric function following the practice of CLIP, given its superior ability to measure the similarity between semantic vectors. Thus, we determine the likelihood of an unlabeled sample belonging to class $c$ as follows:
\begin{align}
    P(y_c|\mathbf{E}^\mathrm{q}, \mathcal{D}^\mathrm{s}_c) = \textstyle{\sum_{\mathbf{E}_c^\mathrm{s} \in \mathcal{D}^\mathrm{s}_c}} \langle\mathbf{E}^\mathrm{q}\cdot \mathbf{E}_c^\mathrm{s}\rangle + \gamma \langle\mathbf{E}^\mathrm{q}\cdot \mathbf{E}_c^\mathrm{TX}\rangle,
\end{align}
where $\mathcal{D}^\mathrm{s}$ is the support set, $\mathbf{E}^\mathrm{s}$ and $\mathbf{E}^\mathrm{q}$ denote the embeddings of a support and query sample,
$\mathbf{E}_c^\mathrm{TX}$ represents the text embedding of \rev{class $c$}%
, and $\gamma$ is a hyperparameter that signifies the weight of label text. Finally, we take the maximum of the computed likelihoods to yield the prediction. %

\section{Dataset and Implementation} \label{sec:impl}
In this section, we introduce the dataset collection and processing, as well as the system implementation of \name.

\subsection{Dataset}
\label{subsec:Dataset}
For the RF modality, we acquire data using a Texas Instruments (TI) IWR1443 Boost mmWave radar~\cite{ti_iwr1443boost}. This radar operates within the 76-81~\!GHz frequency spectrum, offering a bandwidth of 4~\!GHz. It employs a frequency-modulated continuous-wave (FMCW) technique, which transmits a chirp signal that linearly increases in frequency over time. 
The system, upon receiving the reflected signals from the objects, constructs a point cloud. This point cloud aggregates the data collected over a time span of 200~\!ms, and contains information such as point coordinates $(x, y, z)$, Doppler features $d$, and signal intensity $I$. 
\rev{Our dataset for CKD consists of 90,000 video samples (each 200 ms in length), totaling approximately 5 hours in duration.}
Given that the frequencies of most human activities lie within the 0.1-10~\!Hz range~\cite{park2009cadence}, we set the radar sampling rate to 20~\!Hz. After denoising with a constant false alarm rate (CFAR) filter, the resulting point cloud data become $P_i=\left(x_i, y_i, z_i, d_i, I_i\right), 1 \leq i \leq N$, where $N$ denotes the number of points per frame. %

Similarly, we position a Microsoft Kinect V2 RGB camera ~\cite{microsoft_kinect} at the same conditions as the aforementioned mmWave radar. This camera is set to capture images with a resolution of $1920\times1080$ (1080P) and a frame rate of 30~\!Hz. The Kinect V2 captures raw data streams, which are then converted into JPG format to align with the input requirements of the FM. To synchronize these two modalities, which operate at different sampling rates, we initially establish specific start and end actions to assist in preliminary alignment. Subsequently, we select the lower frequency, i.e., the radar frequency, as a reference and identify the temporally closest camera frame for matching, thereby constructing our dataset.

For data acquisition, the pair of radar and camera sensors are positioned in various locations, including being mounted on different desktops, walls, and ceilings. The subjects' heights range from 152 to 186~\!cm, weights from 51 to 109~\!kg, and ages from 10 to 35 years, with an equal distribution of genders. The distance from the sensor to the target ranges from 1 to 15 meters. The dataset is collected across 10 distinct environments: kitchen (KC), living room (LR), bedroom (BR), gym (GM), parking lot (PL), hallway (HW), staircase (SC), park (PK), street (ST), and stadium (SD). \rev{The kitchen, living room, bedroom, and hallway represent limited-space living environments, each furnished with scene-specific items (e.g., different furnitures, hydrants, and ladders). The gym and parking lot are spacious indoor scenes, equipped with fitness equipment and vehicles respectively, and host a modest number of individuals. As outdoor environments, park, street, and stadium are open areas featuring different plants, vehicles, large sports equipment, and pedestrians. The staircase, characterized by its narrow space and complex environment, includes stairs and railings. Collectively, these 10 different environments exhibit unique floor plans and background objects, underscoring the diversity of real-world scenarios.}

Additionally, as elaborated in \S~\ref{subsec:Cross-Modality Distillation}, our dataset is divided into two main parts: everyday spontaneous activities and structured rehabilitation exercises. For the former, approximately 65,000 image-RF data pairs are collected, capturing participants performing activities in accordance with their natural behavior patterns. The latter category encompasses five exercises, each developed in accordance with professional sports rehabilitation guidelines and performed by subjects in compliance with a standardized regimen, ultimately producing approximately 30,000 sample pairs encompassing a broad range of body poses. 

\setcounter{figure}{9}
\begin{figure*}[!b]

	   \captionsetup[subfigure]{justification=centering}
		\centering
		\subfloat[Zero-shot \name.]{
		        \centering
			    \includegraphics[width = 0.24\textwidth]{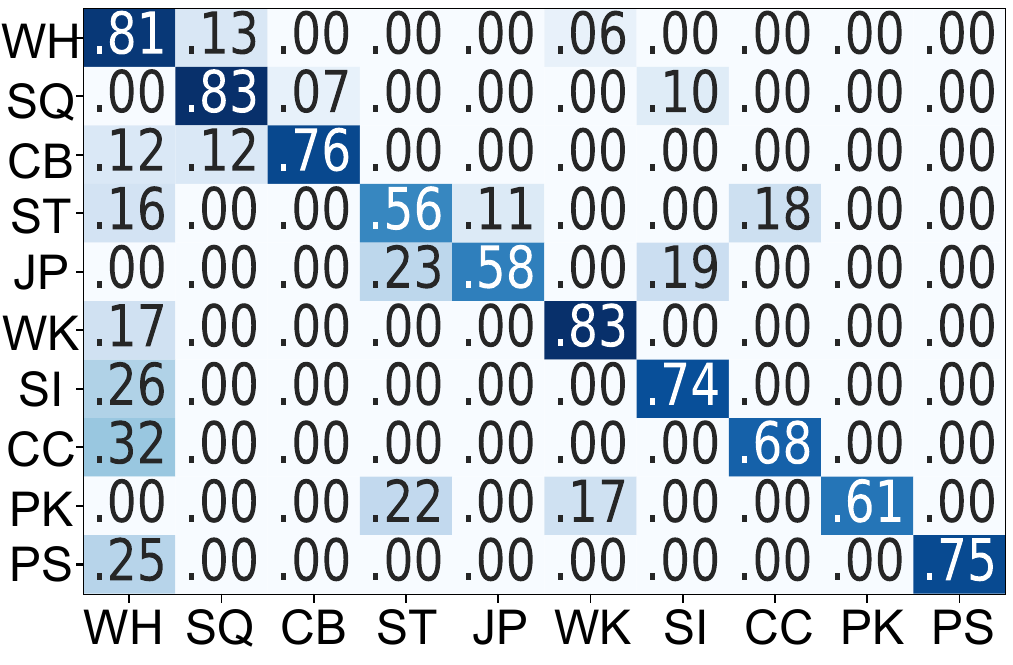}
			\label{subfig:fig_overall_2a}
		}
            \subfloat[1-shot \name.]{
		        \centering
			    \includegraphics[width = 0.24\textwidth]{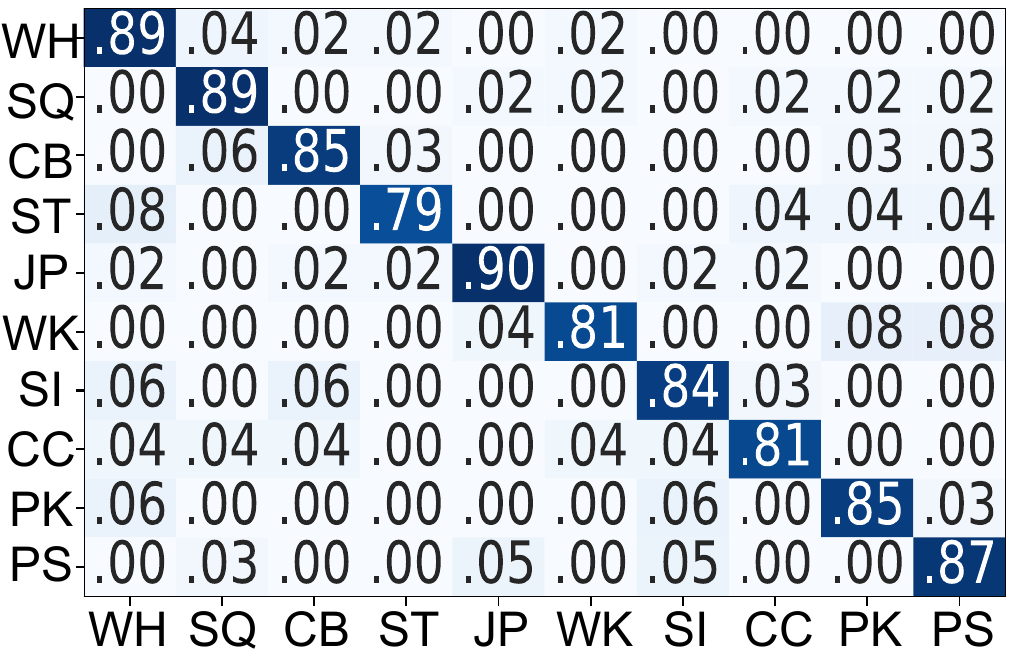}
			\label{subfig:fig_overall_2b}
		}
            \subfloat[3-shot \name.]{
		        \centering
			    \includegraphics[width = 0.24\textwidth]{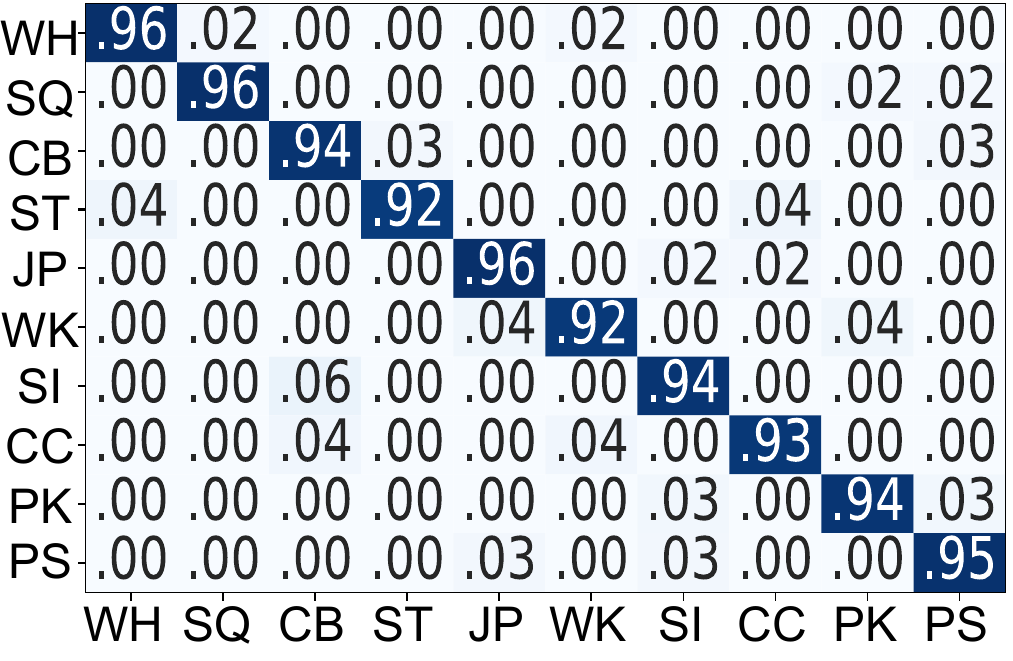}
   
			\label{subfig:fig_overall_2c}
		}
            \subfloat[Zero-shot (10 new classes).]{
		        \centering
			    \includegraphics[width = 0.24\textwidth]{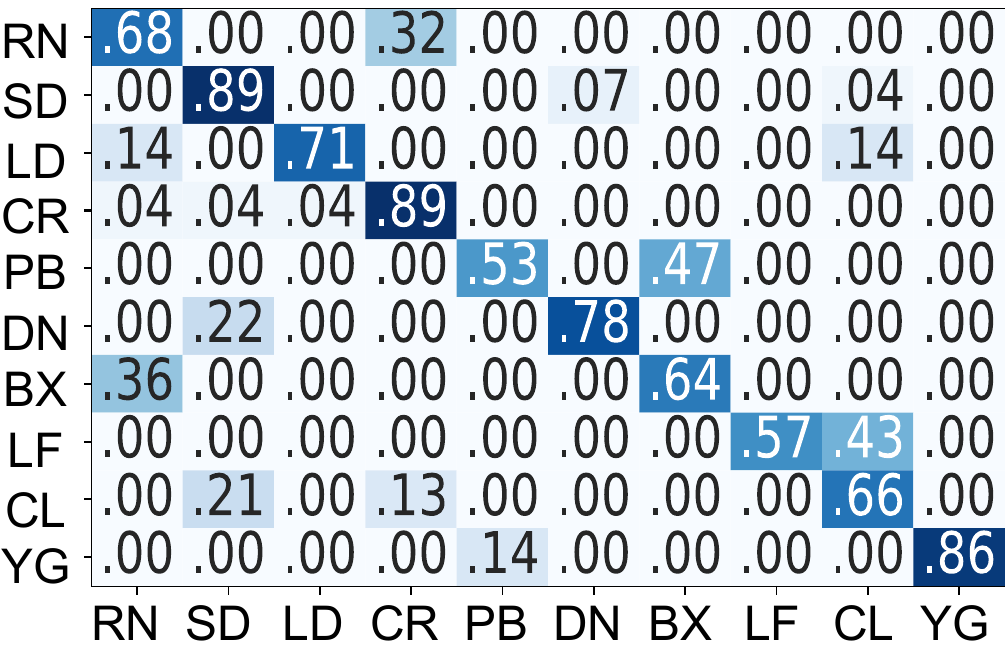}
			\label{subfig:fig_overall_2d}
		}
            
		\caption{Confusion matrices of \name in zero-shot and few-shot scenarios.}
		\label{fig:fig_overall_2}
\end{figure*}

\subsection{System Implementation}
We conduct all experiments, including model training, inference, and saliency map generation, on an NVIDIA TESLA V100 GPU equipped with 16GB of RAM. Regarding software, our framework is built upon Python 3.7 and PyTorch version 2.1.0, which supports CUDA 12.1. Additionally, we employ OpenAI's CLIP as our FM teacher model. The CLIP library, released by OpenAI, facilitates easy integration in Python, providing built-in data preprocessing and a selection of vision encoders. For the RF modality, we develop an mmWave point cloud encoder using PyTorch. The specific configurations are as follows:

\begin{itemize}
    \item  We choose ViT-B/32 in CLIP as our vision encoder and a custom mmWave point cloud encoder, outlined in \S~\ref{ssec:rf_encoder}, featuring 1-d convolutional and linear layers with batch normalization and a 0.3 dropout rate. %
    \item %
    For feature selection, we set a saliency threshold of 0.6, applying Gaussian blur exclusively to regions falling below this threshold using a kernel size of 30.
    \item We employ an Adam optimizer with a learning rate of 0.001 for both cross-modal distillation and few-shot learning, with the latter exploring 1 to 3 shots.
    \item Our CKD dataset consists of 90,000 pairs of image and RF data. The labeled RF dataset has 15,000 samples, and is split into validation and test sets at a 9:1 ratio.
    \item \rev{\name employs continuous, non-overlapping frames for training and testing, instead of random sampling of frames to avoid overfitting caused by neighboring frames.}
    
\end{itemize}

\section{Evaluation} \label{sec:evaluation}
In this section, we report a thorough evaluation on \name in several scenarios and under various parameter settings.

\subsection{Experiment Setup}
\label{subsec:experiment_setup}
To evaluate the performance of \name, we select 3 sets of baselines for comparison. First, 
we compare the \name's rapid adaptation capabilities in RF modality for HAR with limited samples against state-of-the-art (SOTA) meta-learning-based RF models, RF-Net~\cite{ding2020rf} and MetaSense~\cite{gong2019metasense}. Further, we compare the performance of \name against SOTA point-cloud models, PointNet++~\cite{qi2017pointnet++} and Point Transformer~\cite{zhao2021point}. Lastly, to assess \name's performance in unseen environments, we include its teacher model CLIP~\cite{radford2021learning} for comparison.

\begin{itemize}
    \item \textbf{RF-Net} employs a dual-path architecture to discern key RF signal features for HAR and integrates a distance metric network to facilitate few-shot learning.
    \item \textbf{MetaSense} trains on multiple tasks calibrated to individual variances, enabling the model to quickly adapt to new conditions with minimal samples.
    \item \textbf{Point Transformer} introduces a self-attention-based architecture tailored for 3D point cloud analysis that can be used for segmentation and classification tasks.
    \item \textbf{PointNet++ } is an extension of the original PointNet architecture, introducing hierarchical feature extraction to better handle local structures in point clouds.
\end{itemize}
Although \name does not limit the number of HAR classes, we test it on 10 classes for clarity: waving hands $WH$, squatting $SQ$, climbing $CB$, stretching $ST$, jumping $JP$, walking $WK$, sitting $ST$, cycling $CC$, picking $PK$, and pushing $PS$. We also prepare 10 new classes for further evaluation: running $RN$, standing $SD$, lying down $LD$, crawling $CR$, playing ball $PB$, dancing $DN$, boxing $BX$, lifting $LF$, cleaning $CL$, and doing Yoga $YG$.
To gain insights into the model's predictive distribution, we also employ confusion matrices to visually demonstrate the model's performance on each class.
\rev{The experiments strictly follow the IRB approved by our institution.}

\setcounter{figure}{8}
\begin{figure}[t]
    \setlength\abovecaptionskip{6pt}
    \vspace{-1.5ex}
	   \captionsetup[subfigure]{justification=centering}
		\centering
		\subfloat[CLIP.]{
		  \begin{minipage}[b]{0.47\linewidth}
		        \centering
			    \includegraphics[width = 0.96\textwidth]{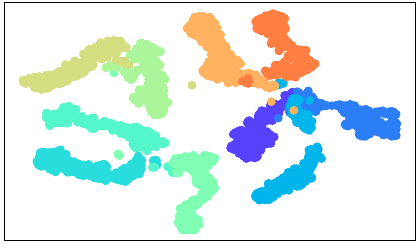}
			\end{minipage}
			\label{subfig:fig_overall_1a}
		}%
		\subfloat[\name.]{
		    \begin{minipage}[b]{0.47\linewidth}
		        \centering
			    \includegraphics[width = 0.96\textwidth]{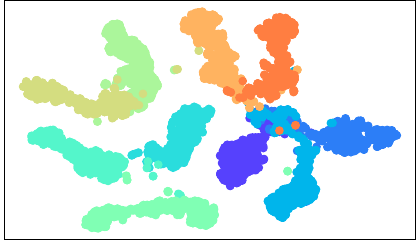}
			\end{minipage}
			\label{subfig:fig_overall_1b}
		}
		\caption{t-SNE plot of embeddings.}
		\label{fig:fig_overall_1}
		
\end{figure}

\subsection{Overall Evaluation of \name}
To evaluate whether \name has acquired CLIP's embedding capability, we first encode image frame-RF sample pairs from our test set into embedding pairs. These 512-dimensional embeddings are then reduced to 2 dimensions for visualization via t-SNE. 
From Figure ~\ref{subfig:fig_overall_1a}, it is evident that the embeddings produced by the CLIP encoder are distinct and well-separated, indicating a high degree of discriminability in the embedding space and a robust capacity for image understanding.
Figure~\ref{subfig:fig_overall_1b} shows that \name's embeddings are separable and closely aligned with the teacher model's, indicating that \name has effectively captured the teacher model's representational power.

In Figure~\ref{fig:fig_overall_2}, we show \name's performance across various zero/few-shot scenarios.
It can be seen that even in the challenging zero-shot context, \name is capable of basic HAR tasks with a notable 72.5\% accuracy. \name also achieves accuracies of 86.0\% and 94.4\% for 1-shot and 3-shot learning. For the 1-shot case, a significant concentration of samples along the confusion matrix diagonal, indicates that \name maintains robust precision and recall for all categories. This level of performance enables accurate HAR task execution. With three labeled samples, the model's accuracy further improves, with the diagonal average approaching 95\%, illustrating a high degree of prediction confidence. Following the few-shot learning phase, we assess \name's performance on 10 new activities mentioned in \S~\ref{subsec:experiment_setup}.
Figure~\ref{subfig:fig_overall_2d} illustrates that the accuracy on new activities aligns with the results in Figure~\ref{subfig:fig_overall_2a}, indicating that the few-shot learning module has a minimal impact on zero-shot performance.

We assess the impact of the number of classes on model accuracy by analyzing both zero-shot and 3-shot performance when the number of classes ranges from 5 to 20, as depicted in Figure~\ref{fig:overall3}. The results reveal a decrement in accuracy as the number of classes increases, with zero-shot learning experiencing a more substantial reduction than 3-shot learning. This trend can be attributed to decreased inter-class distinction and increasing semantic overlap as the number of classes increases, undermining the performance of semantic-driven zero-shot methods. In contrast, the metric-based few-shot classification, which utilizes anchors within the embedding space to enhance decision boundaries, exhibits less performance degradation compared its zero-shot counterpart.

\setcounter{figure}{10}
\begin{figure}[t]
  \centering
  \begin{minipage}{.23\textwidth}
    \centering
    \vspace{0.6ex}
    \includegraphics[width=.99\linewidth]{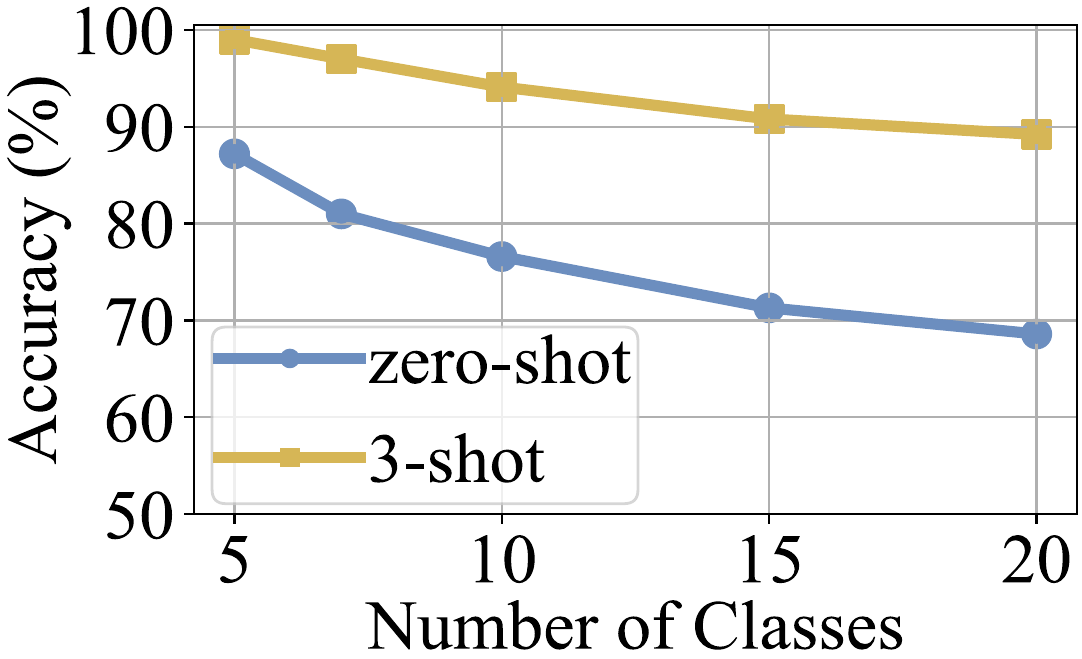}
    \captionof{figure}{Impact of the number of classes.}
    \label{fig:overall3}
  \end{minipage}\hfill
  \begin{minipage}{.23\textwidth}
    \centering
    \vspace*{-1.1ex}
    \includegraphics[width=.99\linewidth]{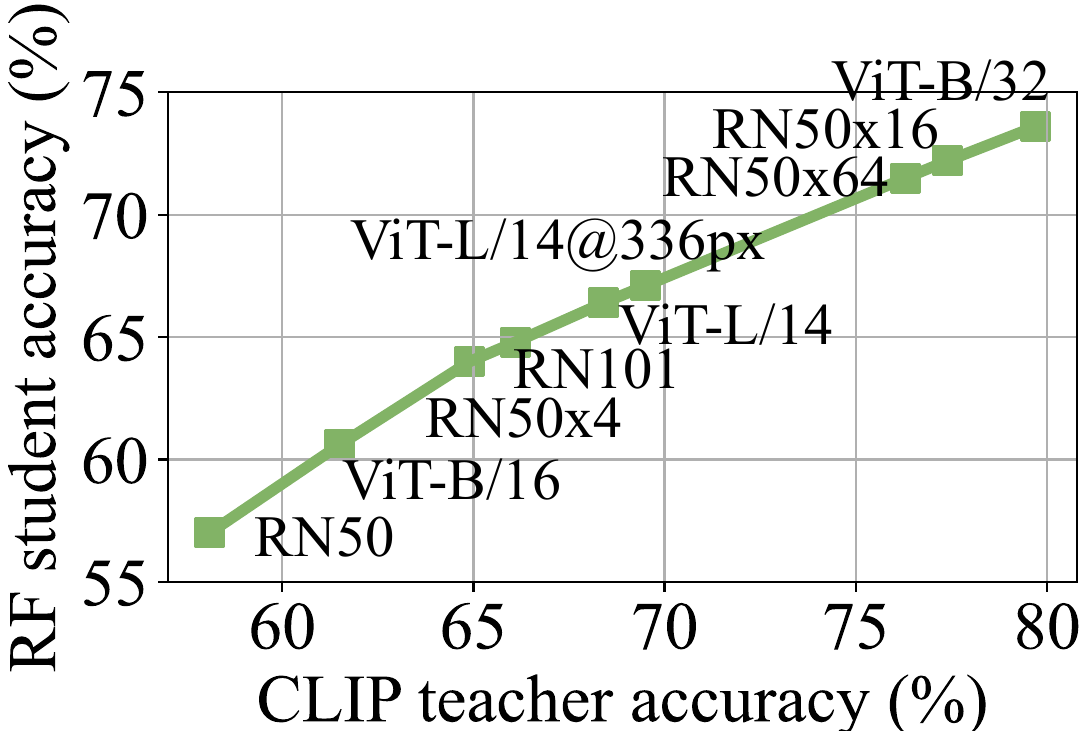}
    \captionof{figure}{Student vs. teacher accuracy.}
    \label{fig:overall4}
  \end{minipage}
  
\end{figure}

Furthermore, we examine the impact of teacher model performance on the effectiveness of the RF student model. As shown in Figure~\ref{fig:overall4}, a stronger teacher model is associated with improved performance of the student model. This results from the teacher's ability to direct the optimization process towards a more efficient trajectory. Notably, 
the student model's size constraints result in decreased performance gains, indicative of an asymptotic trend.
Consequently, ViT-B/32 is chosen as our teacher model backbone due to its superior accuracy of 79.7\% on the zero-shot HAR task, with the corresponding student model also evaluated in the same setting, achieving 73.6\% accuracy.
\rev{Compared with the vision modality, the RF modality shows no significant decline in performance, demonstrating that CKD effectively bridges the modality gap within the embedding space.}

\begin{figure}[b]
    \setlength\abovecaptionskip{6pt}
    \vspace{-6ex}
    
	   \captionsetup[subfigure]{justification=centering}
		\centering
		\subfloat[CKD dataset size.]{
		  \begin{minipage}[b]{0.499\linewidth}
		        \centering
			    \includegraphics[width = 0.96\textwidth]{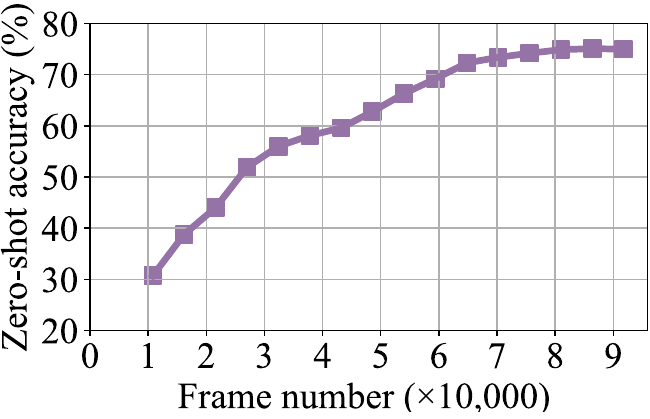}                   
			\end{minipage}
			\label{subfig:overall5a}
		}%
		\subfloat[Model size.]{
		    \begin{minipage}[b]{0.499\linewidth}
		        \centering
			    \includegraphics[width = 0.96\textwidth]{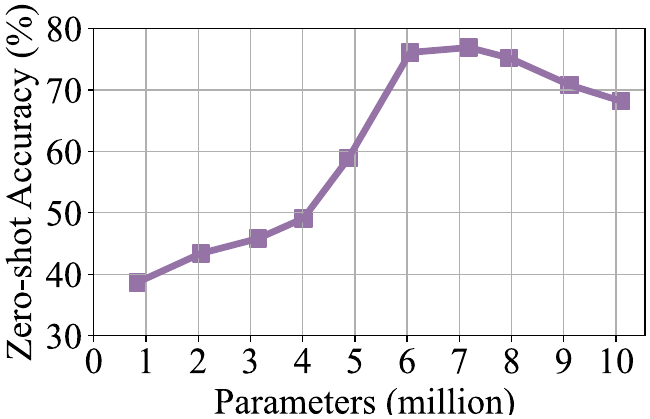}
                    
			\end{minipage}
			\label{subfig:overall5b}
		}
            \vspace{-3ex}
            \subfloat[\rev{Subject-sensor distance}.]{
		    \begin{minipage}[b]{0.499\linewidth}
		        \centering
			    \includegraphics[width = 0.96\textwidth]{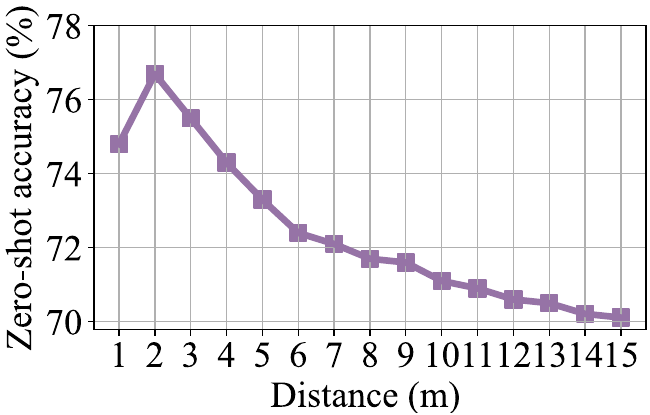}
                    
			\end{minipage}
			\label{subfig:overall5c}
		}
            \subfloat[\rev{Subject-sensor angle}.]{
		    \begin{minipage}[b]{0.499\linewidth}
		        \centering
			    \includegraphics[width = 0.96\textwidth]{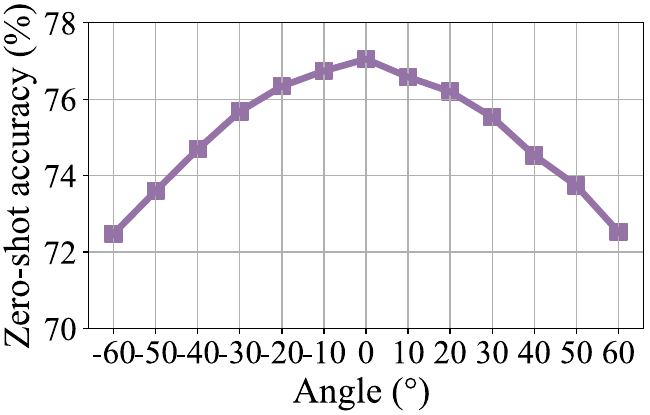}
                    
			\end{minipage}
			\label{subfig:overall5d}
		}

		\caption{Impact of practical factors.}
		\label{fig:overall5}
		
\end{figure}

Next, we investigate the impact of practical factors such as the dataset size for CKD and model complexity on the performance of \name. As depicted in Figure~\ref{subfig:overall5a}, the zero-shot accuracy increases as the number of CKD data samples increases from 10,000 to 90,000, but stops to increase when the number of CKD data reaches 80,000, stabilizing at approximately 75\%. This is close to the 79.7\% accuracy of the teacher model, indicating the efficacy of \name's CKD. We then examine the impact of the number of model parameters, as shown in Figure~\ref{subfig:overall5b}. It can be observed that \name's zero-shot accuracy improves with as the number of parameters increases, reaching a peak of 77\% when the number of parameters reaches 7 million. However, expanding the model further to 10 million parameters leads to overfitting and a notable decline in performance due to the increased model complexity. %

\rev{Finally, we conduct experiments to evaluate the performance of \name when the distance and angle from the subject to the sensor vary. Specifically, the subject performed activities at distances ranging from 1 to 15~\!m and at angles from -60$^\circ$ to 60$^\circ$, which corresponds to the radar's FoV. For each fixed distance, we calculate the average accuracy across all angles; likewise, for each fixed angle, we average the accuracy over all distances. Figure~\ref{subfig:overall5c} demonstrates that the model’s accuracy initially increases as the distance grows, then decreases beyond 2~\!m, peaking at 76.7\%. This optimal performance at 2~\!m is due to the radar's ability to fully capture the subject's body at this distance. At shorter distances, the beam of the radar cannot encompass the entire body; and at larger distances, the signal-to-noise ratio diminishes, lowering the quality of the input data. Despite the degradation, the accuracy remains consistently above 70.2\%, illustrating the strong generalization capability of \name under varying distance conditions. Similarly, Figure~\ref{subfig:overall5d} shows that the model's accuracy decreases as the subject's absolute angle relative to the sensor increases. However, the accuracy remains above 72.9\% across all angles and peaks at 76.7\% when the subject directly faces the sensor (angle of 0$^\circ$). The decline in performance becomes more pronounced near the edges of the FoV due to a sharper drop in radiated power and signal quality. Nonetheless, the performance degradation is modest, with the decrease not exceeding 3.8\%. Collectively, these results demonstrate that \name maintains robust performance, effectively handling variations in both distance and angle, making \name delivers consistent and high accuracy, ensuring sufficient and reliable coverage for HAR.}

\subsection{Superiority of \name}
\subsubsection{Comparison with FM}
We compare \name with FM by assessing their zero-shot capabilities.
\begin{figure}[t]
    \setlength\abovecaptionskip{6pt}
    \vspace{-3ex}
	   \captionsetup[subfigure]{justification=centering}
		\centering
		\subfloat[Zero-shot HAR.]{
		  \begin{minipage}[b]{0.499\linewidth}
		        \centering
			    \includegraphics[width = 0.96\textwidth]{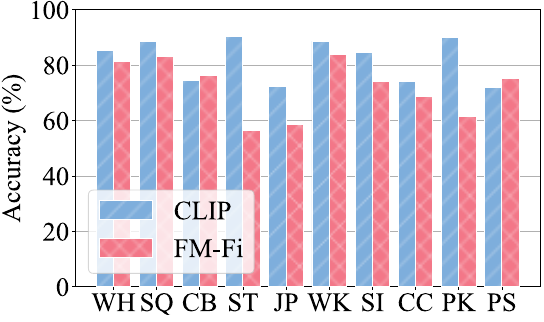}
                    
			\end{minipage}
			\label{subfig:fig13a}
		}
		\subfloat[Resource usage.]{
		    \begin{minipage}[b]{0.499\linewidth}
		        \centering
			    \includegraphics[width = 0.96\textwidth]{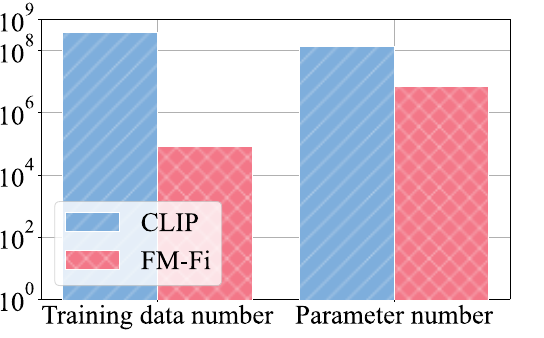}
                    
			\end{minipage}
			\label{subfig:fig13b}
		}
		\caption{Comparison with FM baseline.}
		\label{fig:fig13}
    \vspace{-3ex}
\end{figure}
As shown in Figure~\ref{subfig:fig13a}, the accuracy of \name closely matches that of CLIP across all 10 activity classes, illustrating the overall effectiveness of \name. An interesting phenomenon is that for the two classes of $CB$ and $PS$, the RF-based student model achieves higher accuracy than the FM-based teacher model. The improvement can be attributed to the fact that RF modality might be less susceptible to background image patterns than FM, and the extraneous feature elimination enables CKD to transfer knowledge without irrelevant signals. Additionally, it should be noted that our collected dataset of 90,000 
image-RF pairs is sufficient for CKD. It is also worth mentioning that \name's model, with its 6.9 million parameters, is significantly smaller than CLIP's 140 million parameters, as depicted in Figure~\ref{subfig:fig13b}. 
\rev{Although the model-to-data size ratio of \name exceeds that of typical LLMs, it still achieves strong performance. This distinction can be attributed to two key factors: first, the knowledge distillation paradigm leverages the fact that the teacher model (i.e., CLIP) is trained on an extensive dataset, allowing it to transfer robust and useful representations to the student model. Second, our smaller dataset, which consists of both unstructured data and rehabilitation activity data, is of high quality and highly relevant to the task at hand.}
These observations highlight \name's ability to deliver competitive performance with considerably less data and a more compact architecture.

\subsubsection{Comparison with Few-shot Baselines}
\begin{figure}[t]
    \setlength\abovecaptionskip{6pt}
    
	   \captionsetup[subfigure]{justification=centering}
		\centering
		\subfloat[Few-shot learning.]{
		  \begin{minipage}[b]{0.48\linewidth}
		        \centering
			    \includegraphics[width = \textwidth]{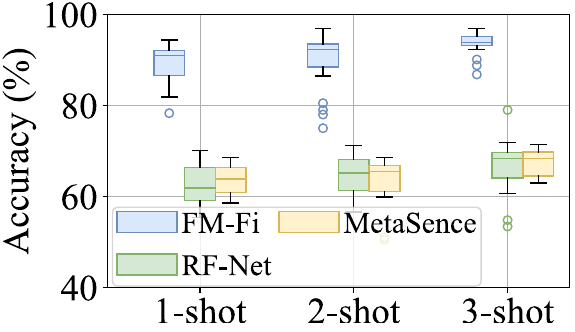}
			\end{minipage}
			\label{subfig:fig11a}
		}
		\subfloat[Supervised learning.]{
		    \begin{minipage}[b]{0.48\linewidth}
		        \centering
			    \includegraphics[width = \textwidth]{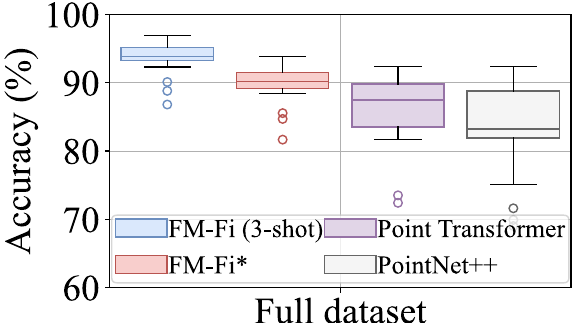}
			\end{minipage}
			\label{subfig:fig11b}
		}
		\caption{Comparison with different baselines.}
		\label{fig:diff_baselines}
		
\end{figure}

We further compare \name with two few-shot baselines MetaSense and RF-Net.  In the experiment, we employ 10-way-$K$-shot learning by sampling $K$ instances from each of 10 classes, creating a shared training set for all models. Figure~\ref{subfig:fig11a} features boxplots that detail the comparative performance of 
them under 1, 2, and 3-shot settings. 
In all three scenarios, \name consistently outperforms the two baselines by a significant margin. Although as the number of samples increases, the median accuracy of \name does not rise as quickly as that of the baselines, it still maintains a lead of at least \rev{23.7\%.} %
Furthermore, the interquartile range (IQR) of \name's accuracy is considerably smaller than that of the baselines, indicating less variability across multiple experiments. %

\begin{figure}[b]
    \setlength\abovecaptionskip{6pt}
    
	   \captionsetup[subfigure]{justification=centering}
		\centering
            \vspace{-3ex}
		\subfloat[Class-wise accuracy.]{

		  \begin{minipage}[b]{0.50\linewidth}
		        \centering
			    \includegraphics[width = \textwidth]{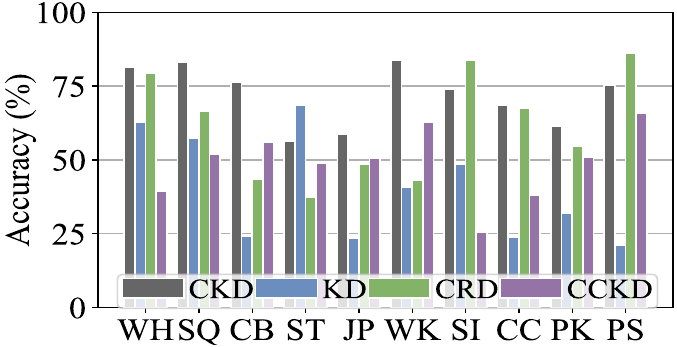}
                \vspace{-1ex}    
			\end{minipage}
			\label{subfig:ckd_eval1}
		}
		\subfloat[Number of classes.]{
		    \begin{minipage}[b]{0.49\linewidth}
		        \centering
			    \includegraphics[width = \textwidth]{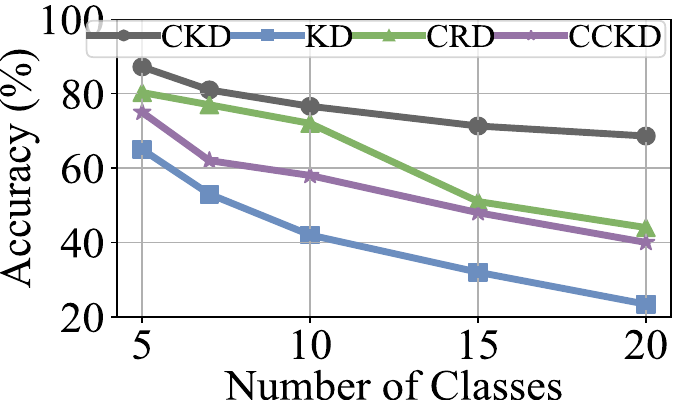}
                    
			\end{minipage}
			\label{subfig:ckd_eval2}
		}
		\caption{Generalization comparison.}
		\label{fig:ckd_eval}
		
\end{figure}

\subsubsection{Comparison with Supervised Baselines on a Larger Dataset}

We further compare \name with PointNet++ and Point Transformer. These models are trained on an expanded dataset (including 50,000 labeled RF samples) without CKD. For ease of comparison, we introduce an additional baseline model termed FM-Fi*, which utilizes the same RF encoder as \name (with an ensuing multilayer perception for converting the embedding to classification result). FM-Fi* is also trained on the same 50,000-sample dataset without CKD.
Using only 0.1\% of the labeled data compared to the other three models, 3-shot \name not only demonstrates superior accuracy but also greater stability in performance.
These results highlight the efficacy of CKD in learning robust representations while significantly decreasing the dependency on annotated RF data. Furthermore, among the three fully supervised models, FM-Fi* exhibits notably better performance than the other two models that are specifically designed for point clouds. The superior performance of FM-Fi* is due to its RF encoder which effectively integrates point cloud coordinates with Doppler features and signal intensity, thus utilizing the complete range of information available in RF data. %

\subsection{CKD Evaluation} \label{subsec:CKD}
We further compare CKD with KD, as well as contrastive representation distillation (CRD)~\cite{tian2019contrastive}, and correlation congruence for knowledge distillation (CCKD)~\cite{peng2019correlation}. First, we compare their performance on a 10-class zero-shot HAR task, as illustrated in Figure~\ref{subfig:ckd_eval1}. 
We observe that CKD achieves the highest accuracy in 7 out of 10 classes, only trailing the best method by less than \rev{9.8\% }%
in the rest 3 classes. Figure~\ref{subfig:ckd_eval2} further examines the impact of the number of activity classes. It can be observed that when the number of classes is 5, CKD leads other methods by a small margin less than \rev{18.5\%.}%
As the number of classes increases to 20, CKD exhibits the smallest drop in accuracy, while the accuracies of all other methods drop by \rev{more than 37.3\%.}%

To understand CKD’s superiority, we analyze the relationship between the \name's accuracy and the extent of interdependency information transfer. We employ the mean differences in the correlation matrices of image/RF embeddings to quantify interdependency transfer. By varying $\tau$ in the CKD loss, the correlation differences can be adjusted. Our findings shown in Figure~\ref{fig:correlation} demonstrate that the correlation difference negatively impacts \name's accuracy ($\tau=10$ yields the best performance). This trend validates \name’s principle: preserving the interdependency information among the embedding elements is crucial for HAR. In contrast, the inferior results of alternative approaches (indicated by markers near the curve) can be attributed to their pronounced correlation differences, which correspond to a diminished efficacy in the transfer of interdependency knowledge.

\setcounter{figure}{16}
\begin{figure}[t]
  \centering
  \begin{minipage}{.23\textwidth}
    \vspace{3ex}
    \centering
    \includegraphics[width=.99\linewidth]{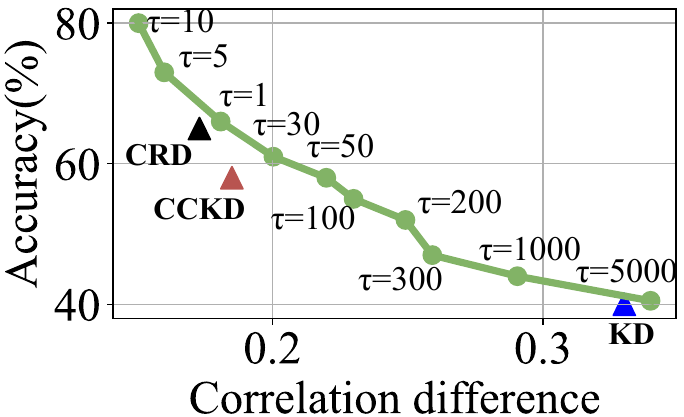}
    \vspace{-5ex}
    \captionof{figure}{Impact of interdependency transfer.}
    \label{fig:correlation}
    
  \end{minipage}\hfill
  \begin{minipage}{.23\textwidth}
    \vspace{2.5ex}
    \centering
    \includegraphics[width=.99\linewidth]{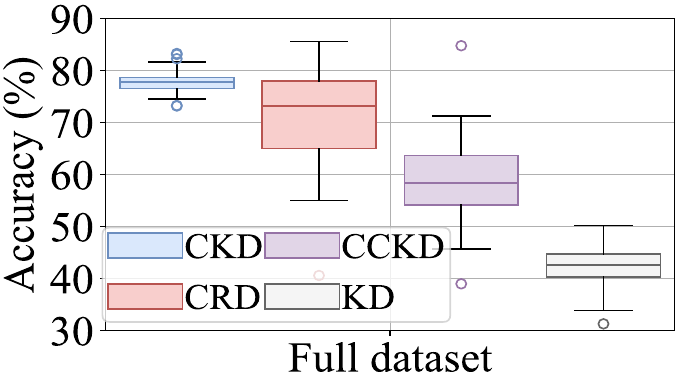}
    \vspace{-2.7ex}
    \captionof{figure}{Comparisons on accuracy and stability.}
    \label{fig:stability}
  \end{minipage}
  \vspace{-3ex}
\end{figure}

Finally, we conduct 50 training rounds (number of classes set to 10), and perform statistical analysis of the accuracies of various distillation methods, and show the results in Figure~\ref{fig:stability}. It can be seen that CKD exhibits the highest median accuracy and narrowest IQR. In contrast, CRD, CCKD, and KD demonstrate lower accuracies and larger IQR. Notably, CRD shows the highest variability in accuracies, which can be attributed to the instability inherent in its learning-based critic model used for similarity assessment. 
CCKD’s approach, which prioritizes alignment of instance distributions between image and RF embeddings without addressing the interdependencies among elements, results in suboptimal performance. Similarly, KD’s performance is compromised due to its inability to manage the interdependencies within the embeddings' elements.
\rev{In summary, CKD's advantages arise from: a greater emphasis on the interdependencies of embedding elements compared to CCKD and KD, which transfers critical information to enhance performance; and using cosine similarity instead of a critic model, as in CRD, which reduces model complexity and increases robustness.}

\subsection{Feature Elimination Evaluation}
\subsubsection{Image Modality}
\begin{figure}[t]
    \setlength\abovecaptionskip{6pt}
    
	   \captionsetup[subfigure]{justification=centering}
		\centering
            \subfloat[Example segmentation methods and activations.]{
            
		  \begin{minipage}[b]{0.44\linewidth}
		        \centering
			    \includegraphics[width = 0.99\textwidth]{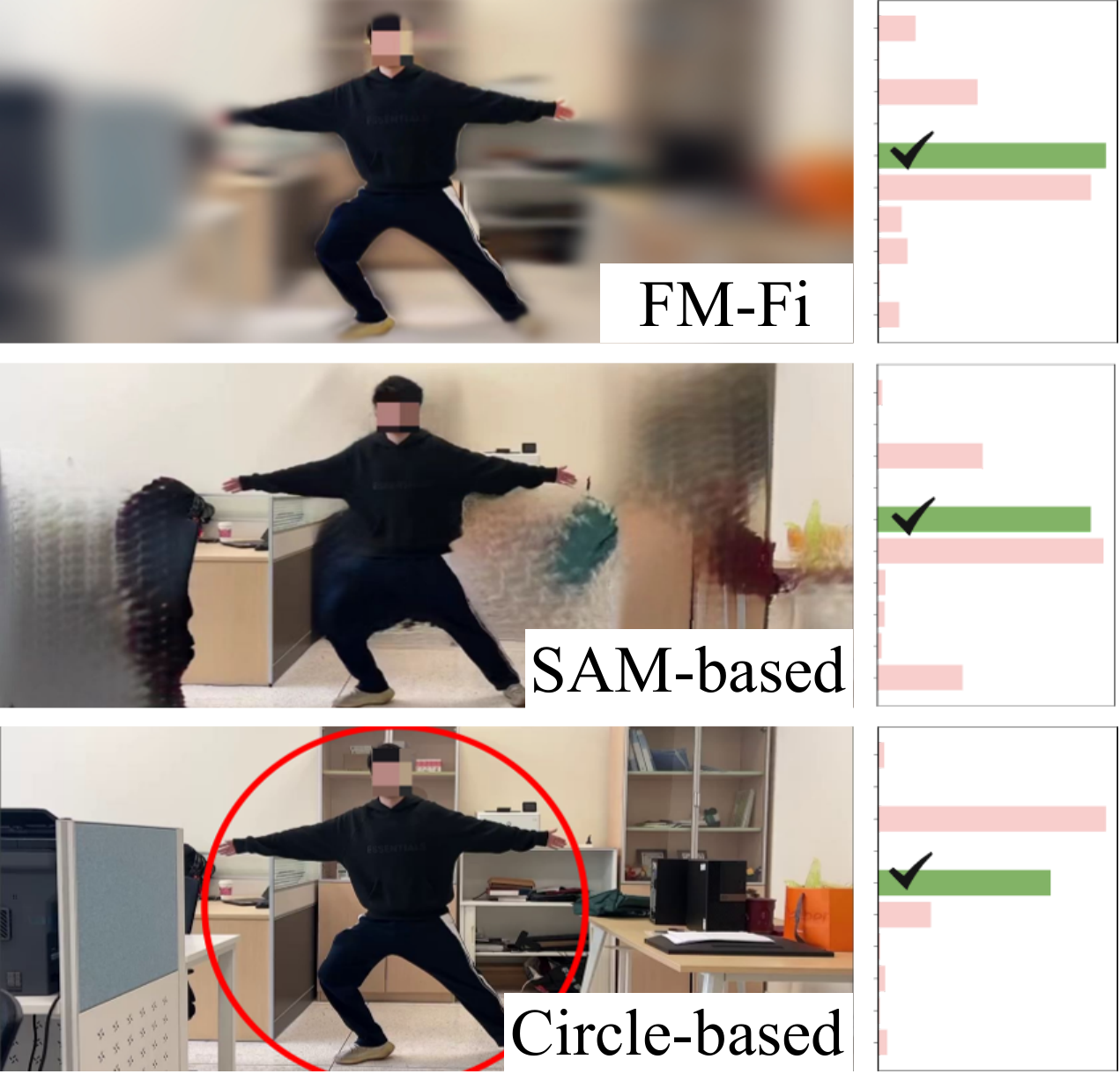}
                \vspace{-0.7ex}    
			\end{minipage}
			\label{subfig:feature_elimination1d}
		}%
            \subfloat[Probability distributions of cosine similarity.]{
		  \begin{minipage}[b]{0.5\linewidth}
		        \centering
			    \includegraphics[width = 0.99\textwidth]{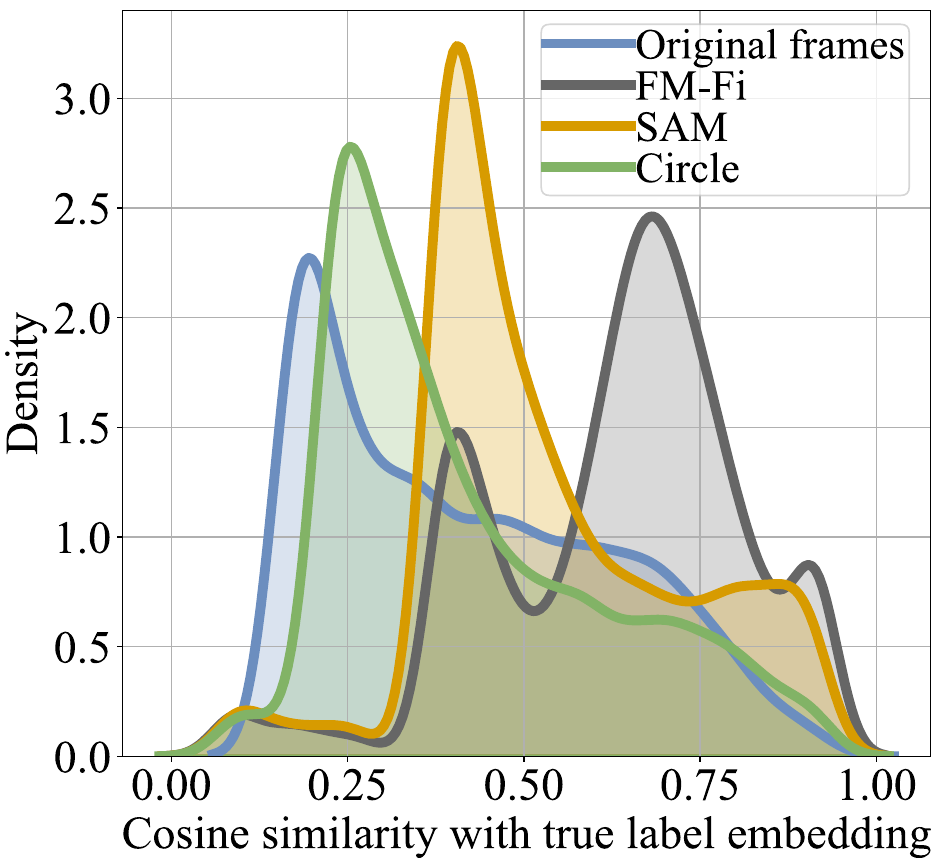}
                    
			\end{minipage}
			\label{subfig:feature_elimination1e}
		}
		\caption{Performance comparison of different image feature elimination methods.}
		\label{fig:feature_elimination1}
    \vspace{-3.5ex}		
\end{figure}

In addition to a saliency map-based feature elimination method, Segment-Anything (SAM)~\cite{kirillov2023segment} and circle prompting~\cite{shtedritski2023does} can also be used to enhance human focus in images. SAM %
\rev{is the SOTA segmentation algorithm and is shown to be superior for the HAR task;}
whereas circle prompting emphasizes crucial information within an image through circular markings made by a human annotator.
Figure~\ref{subfig:feature_elimination1d} depicts the images processed with our saliency map-based method, SAM-based method, and circle-based method, respectively, along with the predictions made by CLIP. The ground truth classes have been highlighted by green for reference. It is evident that due to the extraneous background features, CLIP fails to correctly classify the outcomes from the other two prompting methods. In contrast, our approach effectively mitigates extraneous features, making CLIP solely focus on the human subject, thus enabling more refined classification. The probability distributions of the similarity between true and predicted embedding in Figure~\ref{subfig:feature_elimination1e} further prove the advantage of \name's extraneous feature elimination module. 

\begin{figure}[b]
    \setlength\abovecaptionskip{6pt}
    \vspace{-5ex}
	   \captionsetup[subfigure]{justification=centering}
		\centering
		\subfloat[Image modality.]{
		  \begin{minipage}[b]{0.52\linewidth}
		        \centering
			    \includegraphics[width = 0.96\textwidth]{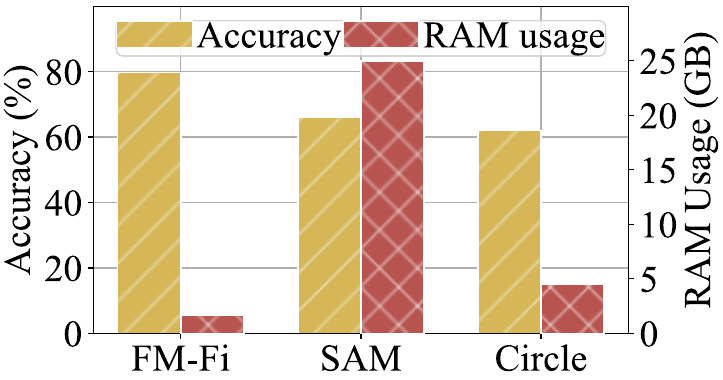}
                    
			\end{minipage}
			\label{subfig:feature_elimination2a}
		}\hspace{-2ex}
		\subfloat[RF modality.]{
		    \begin{minipage}[b]{0.46\linewidth}
		        \centering
			    \includegraphics[width = 0.96\textwidth]{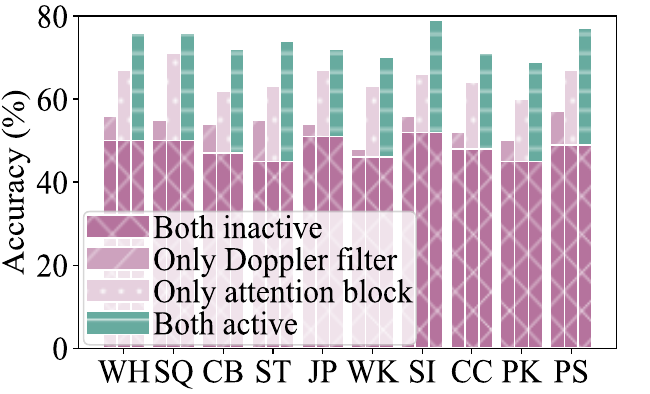}
                    
			\end{minipage}
			\label{subfig:feature_elimination2b}
		}
            
		\caption{Comparison of feature elimination methods.}
		\label{fig:feature_elimination2}
		
\end{figure}

Among the three methods, \name is the only one capable of automatic background removal. This is attributed to \name's ability to autonomously identify people within images through semantic input. In contrast, the SAM-based method requires manual selection post-segmentation to remove background elements, while the circle-based approach relies entirely on manual annotation of significant objects.
We have also compared the resource demands of these methods, noting that manual costs are challenging to assess directly. Therefore, we translate the manual annotation task in the circle-based method into an automated segmentation and highlight-filling task. Figure~\ref{subfig:feature_elimination2a} shows that our approach not only leads in performance but also in reduces resource utilization, thus underscoring the superiority of the saliency map-based feature elimination module of \name.

\subsubsection{RF Modality}
We also evaluate the feature elimination module of the RF modality, which consists of two components: a Doppler-based object filter and an attention block.
To validate the efficacy of the proposed module, we conduct ablation studies by removing the Doppler filter and attention block. As such, the experiment involves the following 4 configurations: both parts inactive, Doppler filter inactive, attention block inactive, and both parts active. We select 10-class classification for evaluation. Figure~\ref{subfig:feature_elimination2b} presents the findings, where we plot the effect of incorporating various feature elimination components on the accuracy of zero-shot classification.

Not surprisingly, when both modules are inactive, the accuracy is the worst and only reaches \rev{less than 52.3\%.}%
The result also shows that the Doppler filter, upon excluding objects with zero velocity, has a positive but limited effect on the model. The attention block, with its automatic selection of important points, offers a more effective improvement to overall performance. The combined effect of both parts surpasses that of each individual component. This synergy can be attributed to the Doppler filter's introduction of physical priors that enhance the subsequent decision-making process of the attention block, thereby underscoring the effectiveness and necessity of our module.

\subsection{Generalization Capability}
\begin{figure}[t]
    \setlength\abovecaptionskip{6pt}
    
	   \captionsetup[subfigure]{justification=centering}
		\centering
		\subfloat[Different scenes.]{
		  \begin{minipage}[b]{0.49\linewidth}
		        \centering
			    \includegraphics[width = 0.96\textwidth]{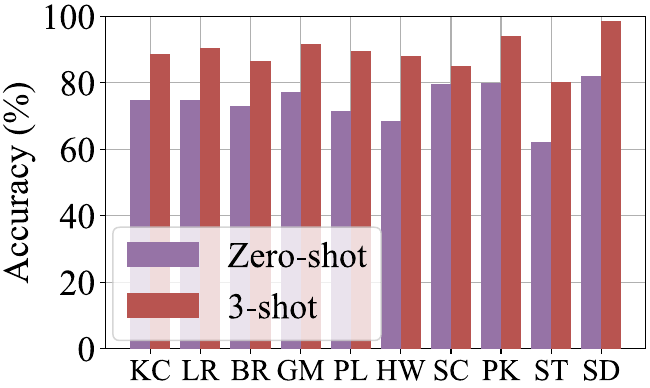}
                    
			\end{minipage}
			\label{subfig:generalization_a}
		}\hspace{-2ex}
		\subfloat[Different subjects.]{
		    \begin{minipage}[b]{0.49\linewidth}
		        \centering
			    \includegraphics[width = 0.96\textwidth]{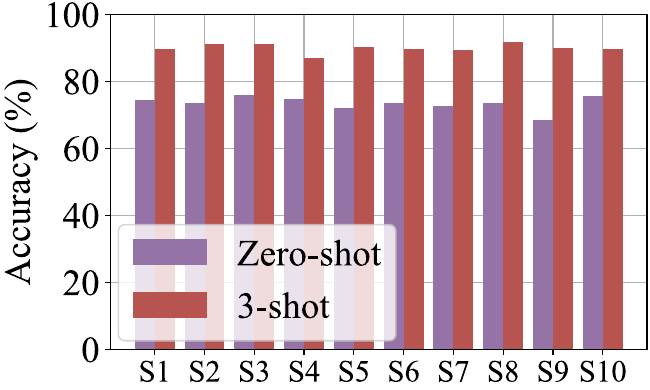}
                    
			\end{minipage}
			\label{subfig:generalization_b}
		}
            
		\caption{Generalization across diverse settings.}
		\label{fig:generalization}
    \vspace{-3ex}		
\end{figure}

To evaluate the generalization capabilities of \name, we conduct tests on the 10 subjects ($S1 - S10$) and 10 environments mentioned in \S~\ref{subsec:Dataset}. 
\rev{Specifically, we adopt a leave-one-out strategy for 3-shot testing, where we train on data from 9 environments or subjects and test on the remaining one; for zero-shot testing, we directly conduct tests without additional training.}
We conduct both zero-shot and 3-shot tests
\rev{in 100 settings (10 environments $\times$ 10 subjects) and the results shown in Figure~\ref{fig:generalization} are obtained by averaging across either environments or subjects.}

Overall, Figure~\ref{subfig:generalization_a} shows that performance tends to be better in outdoor scenes due to factors such as better lighting, open space, less background features, and reduced occlusion. However, street scenes yield poorer results because of the interference from rapidly moving background objects such as cars and pedestrians, which can disrupt RF signals. In contrast, for primary RF-based HAR scenarios, especially in domestic settings, \name maintains performance levels consistent with previous tests, demonstrating exceptional capabilities. Moreover, physiological parameters such as age and height of participants do not affect the performance of \name, as evidenced by Figure~\ref{subfig:generalization_b}, which illustrates that our model maintains \rev{at least 68.5\% }%
zero-shot accuracy and \rev{89.8\% }%
3-shot accuracy across all subjects, demonstrating robust generalization capabilities.

\subsection{Hyper-parameter Searching}
\subsubsection{Feature Elimination Threshold}
\begin{figure}[t]
    \vspace{-2ex}
	   \captionsetup[subfigure]{justification=centering}
		\centering
		\subfloat[Image modality.]{
		  \begin{minipage}[b]{0.48\linewidth}
		        \centering
			    \includegraphics[width = 0.96\textwidth]{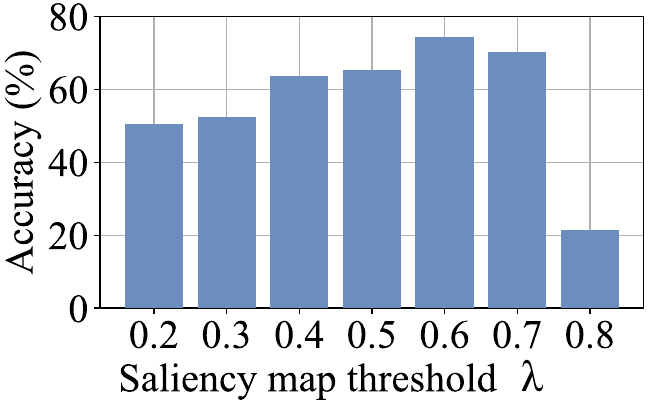}
			\end{minipage}
			\label{subfig:hyper1a}
		}%
		\subfloat[RF Modality.]{
		    \begin{minipage}[b]{0.48\linewidth}
		        \centering
			    \includegraphics[width = 0.96\textwidth]{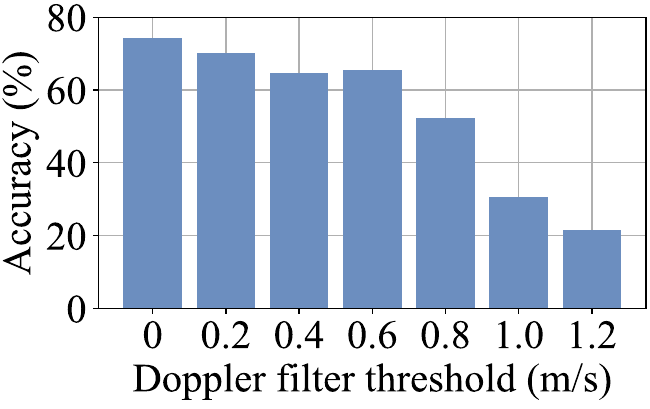}
			\end{minipage}
			\label{subfig:hyper1b}
		}
            \vspace{-2ex}
		\caption{Feature elimination thresholds.}
		\label{fig:hyper1}
    \vspace{-3ex}
		
\end{figure}

According to \S ~\ref{subsubsec:feature elimination image}, the threshold $\lambda$ is a scalar within the [0,1] range, determining the lower bound normalized score for pixels exempt from blur transformation. On one hand, a small $\lambda$ preserves the original image content, but fails to efficiently eliminate background noise. On the other hand, a high $\lambda$ 
value risks removing critical image features, depriving the model of meaningful input and thereby reducing the discriminability of the generated cross-modal supervision signal.
To search for the optimal value of $\lambda$, we evaluate the zero-shot performance of \name at different $\lambda$ values from 0.2 to 0.8. 
One may readily observe in Figure~\ref{subfig:hyper1a} that as $\lambda$ initially increases, \name reaches the best performance at the optimal threshold $\lambda=0.6$. Any $\lambda$ greater than 0.6 may cause the saliency mask to erode the human figure, adversely impacting HAR performance. Consequently, as $\lambda$ surpasses 0.6, there is a significant decline in accuracy from \rev{76.2\% to 21.6\%.}%

We further study the impact of velocity thresholds in extraneous feature elimination for the RF modality. Instead of only removing the zero-velocity component as in \S~\ref{subsubsec:RF Modality}, we set the velocity filtering thresholds from 0 to 1.2~\!m/s, and show the relationship between the model's accuracy and the threshold in Figure~\ref{subfig:hyper1b}.
It is observed that the model performs the best when the Doppler threshold is set to 0, which corresponds to the removal of static background. Increasing the Doppler threshold may inadvertently filter out some moving background clutter; however, it might also eliminate information pertinent to human activities, leading to a decline in model performance. When the Doppler threshold reaches 0.8~\!m/s, a significant portion of human activity information is lost, resulting in poor model performance. Based on these experiment observations, a threshold of 0 is selected to preserve all information of moving objects while excluding static background features, leaving the subsequent attention module to make further selections.

\begin{figure}[b]

    \setlength\abovecaptionskip{6pt}
    \vspace{-5ex}
    
	   \captionsetup[subfigure]{justification=centering}
		\centering
		\subfloat[Preliminary search.]{
		  \begin{minipage}[b]{0.47\linewidth}
		        \centering
			    \includegraphics[width = 0.96\textwidth]{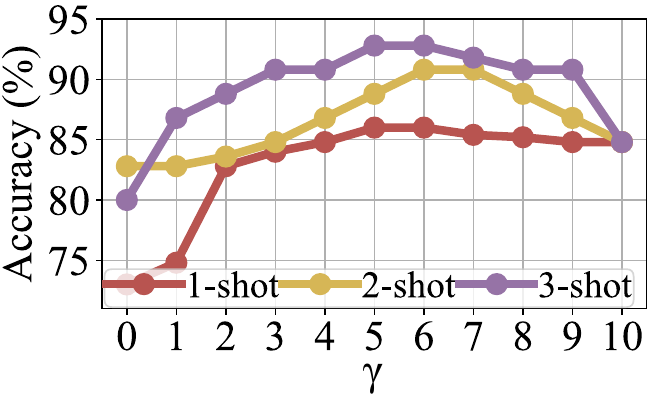}
			\end{minipage}
			\label{subfig:hyper2a}
		}
		\subfloat[Detailed search.]{
		    \begin{minipage}[b]{0.47\linewidth}
		        \centering
			    \includegraphics[width = 0.96\textwidth]{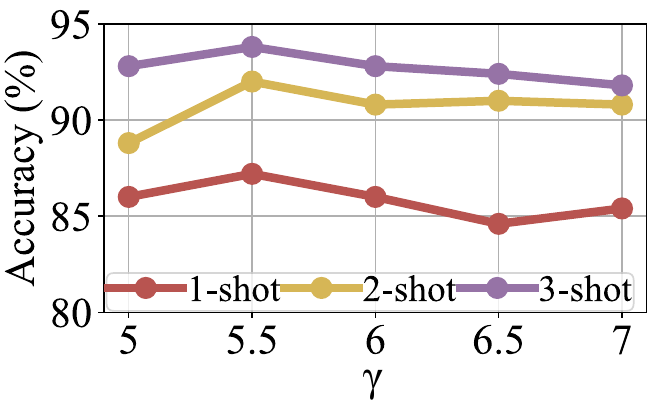}
			\end{minipage}
			\label{subfig:hyper2b}
		}
            
		\caption{Impact of the weight of label text.}
		\label{fig:hyper2}
		
\end{figure}

\subsubsection{Weight of Label Text in Few-shot Learning}
We evaluate the impact of varying weights of label text $\gamma$ on \name's performance across 1-shot, 2-shot, and 3-shot learning scenarios. Initial assessments are conducted with integer values of $\gamma$ in the range 0 to 10, with results depicted in Figure~\ref{subfig:hyper2a}. We observe that for small values, accuracy across all scenarios increased with $\gamma$, suggesting effective semantic information extraction from the RF modality by \name. As $\gamma$ increases, performance across the three scenarios tends to converge due to the text embedding becoming the dominant factor. Such convergence results in performance degradation, approaching zero-shot levels as $\gamma$ further increases. We aim to identify the best performance point $\gamma=5$ to $\gamma=7$. As Figure~\ref{subfig:hyper2b} demonstrates, the peak performance is obtained at $\gamma = 5.5$. Consequently, we adopt a $\gamma$ value of 5.5 as the weight of text label in few-shot learning.

\vspace{-0ex}
\section{Related Work and Discussion} \label{sec:discussion}
Though RF-HAR literature covers enhancing generalizability~\cite{widar,jiang2018towards,ding2020rf, gong2019metasense,gao2022towards,meneghello2022sharp,chen2021rf}, improving the efficient utilization of scarce labeled data~\cite{ouyang2021clusterfl,li2021semisupervised}, and refining model architectures~\cite{chen2021rf,chakraborty2022diat,li2023human}, prominent RF-HAR proposals have prioritized studies on generalizability. In particular, Widar3.0~\cite{widar} introduces a domain-independent and signal-level feature, termed BVP, to enable generalizability.
Another study~\cite{jiang2018towards} applies adversarial domain adaptation techniques~\cite{ganin2015unsupervised,ganin2016domain} to generalize across varying scenarios. RF-Net~\cite{ding2020rf} adopt metric-based meta-learning achieve fast adaptation of its base networks in diverse environments. %

The emergence of FMs has brought new potentials in RF sensing in general, catering the need for more models capable of capturing rich information. 
Therefore, \name sets itself apart from prior RF-HAR solutions by not limiting itself to HAR, because it inherits the broad recognition capability of FM. In fact, we would expect 
\name to be able to support other sensing tasks~\cite{hu2024m,zheng2023autofed,zhang2022quantifying,chen2023wider,zheng2020v2ifi,zhang2022can,10339891} including gesture detection~\cite{liu2019real,melgarejo2014leveraging,zhang2023ochid}, gait recognition~\cite{cao2022cross,wang2019ev,sun2020wearable}, and even vibration monitoring~\cite{adib2015smart,xie2022deepvs,zhang2023pi,chen2021octopus,chen2021movi}, by
modifying the target of interest; we plan to explore \name's potential beyond HAR in future work.
Currently, it is still an open question if one can claim open-set capability for FM-enabled HAR~\cite{ren2022delving}. Also, whether \name may completely inherit the knowledge of FM 
(obtained from massive datasets encompassing a broad spectrum of activities) needs further studies.
\rev{Furthermore, the question of how to compress the RF model by quantization when transferring knowledge from the FM~\cite{cai2024self} is also of practical significance.} 
As \name pioneers in the knowledge transferring from FM to RF-HAR, we leave these uncertainties to future exploration.

\vspace{-0ex}
\section{Conclusion} \label{sec: conclusion}
Taking a significant stride in advancing HAR, we have introduced \name, which harnesses the interpretative power of FMs to facilitate cross-modal RF-HAR. By employing CKD and extraneous feature elimination, the innovative RF encoder in \name\ effectively assimilates the semantic embedding derived from FMs. This enables precise mapping of RF data for efficient zero/few-shot HAR applications, addressing the critical challenge of data scarcity in RF-HAR. Our thorough experiment analysis across diverse and complex scenarios confirms \name's superiority over conventional baselines. This research not only demonstrates the effectiveness of our approach but also lays the groundwork for further advancements in RF-HAR, while aiming for broader 
RF sensing tasks in practical settings.

\vspace{0ex}
\section*{Acknowledgments}
The study is supported by Shenzhen Science and Technology Program (No. 20231120215201001) and the research start-up grant from the Southern University of Science and Technology, for which Tianyue Zheng expresses sincere gratitude. We are also grateful to the anonymous reviewers for their  constructive comments. As a side note, one of the authors, Yanbing Yang, receives funding from National Natural Science Foundation of China (62272329).

\vfill\eject
\balance
\bibliographystyle{ACM-Reference-Format}
\bibliography{main}


\begin{thebibliography}{77}


\ifx \showCODEN    \undefined \def \showCODEN     #1{\unskip}     \fi
\ifx \showDOI      \undefined \def \showDOI       #1{#1}\fi
\ifx \showISBNx    \undefined \def \showISBNx     #1{\unskip}     \fi
\ifx \showISBNxiii \undefined \def \showISBNxiii  #1{\unskip}     \fi
\ifx \showISSN     \undefined \def \showISSN      #1{\unskip}     \fi
\ifx \showLCCN     \undefined \def \showLCCN      #1{\unskip}     \fi
\ifx \shownote     \undefined \def \shownote      #1{#1}          \fi
\ifx \showarticletitle \undefined \def \showarticletitle #1{#1}   \fi
\ifx \showURL      \undefined \def \showURL       {\relax}        \fi
\providecommand\bibfield[2]{#2}
\providecommand\bibinfo[2]{#2}
\providecommand\natexlab[1]{#1}
\providecommand\showeprint[2][]{arXiv:#2}

\bibitem[Adib et~al\mbox{.}(2015)]%
        {adib2015smart}
\bibfield{author}{\bibinfo{person}{Fadel Adib}, \bibinfo{person}{Hongzi Mao}, \bibinfo{person}{Zachary Kabelac}, \bibinfo{person}{Dina Katabi}, {and} \bibinfo{person}{Robert~C Miller}.} \bibinfo{year}{2015}\natexlab{}.
\newblock \showarticletitle{{Smart Homes that Monitor Breathing and Heart Rate}}. In \bibinfo{booktitle}{\emph{Proc. of the 33rd ACM CHI}}. \bibinfo{pages}{837--846}.
\newblock


\bibitem[Bansal et~al\mbox{.}(2020)]%
        {bansal2020pointillism}
\bibfield{author}{\bibinfo{person}{Kshitiz Bansal}, \bibinfo{person}{Keshav Rungta}, \bibinfo{person}{Siyuan Zhu}, {and} \bibinfo{person}{Dinesh Bharadia}.} \bibinfo{year}{2020}\natexlab{}.
\newblock \showarticletitle{{Pointillism: Accurate 3D Bounding Box Estimation with Multi-Radars}}. In \bibinfo{booktitle}{\emph{Proc. of the 18th ACM SenSys}}. \bibinfo{pages}{340--353}.
\newblock


\bibitem[Bi et~al\mbox{.}(2019)]%
        {bi2019familylog}
\bibfield{author}{\bibinfo{person}{Chongguang Bi}, \bibinfo{person}{Guoliang Xing}, \bibinfo{person}{Tian Hao}, \bibinfo{person}{Jina Huh-Yoo}, \bibinfo{person}{Wei Peng}, \bibinfo{person}{Mengyan Ma}, {and} \bibinfo{person}{Xiangmao Chang}.} \bibinfo{year}{2019}\natexlab{}.
\newblock \showarticletitle{{FamilyLog: Monitoring Family Mealtime Activities by Mobile Devices}}.
\newblock \bibinfo{journal}{\emph{IEEE Transactions on Mobile Computing}} \bibinfo{volume}{19}, \bibinfo{number}{8} (\bibinfo{year}{2019}), \bibinfo{pages}{1818--1830}.
\newblock


\bibitem[Boroushaki et~al\mbox{.}(2021)]%
        {boroushaki2021robotic}
\bibfield{author}{\bibinfo{person}{Tara Boroushaki}, \bibinfo{person}{Junshan Leng}, \bibinfo{person}{Ian Clester}, \bibinfo{person}{Alberto Rodriguez}, {and} \bibinfo{person}{Fadel Adib}.} \bibinfo{year}{2021}\natexlab{}.
\newblock \showarticletitle{{Robotic Grasping of Fully-Occluded Objects using RF Perception}}. In \bibinfo{booktitle}{\emph{Proc. of IEEE ICRA}}. IEEE, \bibinfo{pages}{923--929}.
\newblock


\bibitem[Brown et~al\mbox{.}(2020)]%
        {brown2020language}
\bibfield{author}{\bibinfo{person}{Tom Brown}, \bibinfo{person}{Benjamin Mann}, \bibinfo{person}{Nick Ryder}, \bibinfo{person}{Melanie Subbiah}, \bibinfo{person}{Jared~D Kaplan}, \bibinfo{person}{Prafulla Dhariwal}, \bibinfo{person}{Arvind Neelakantan}, \bibinfo{person}{Pranav Shyam}, \bibinfo{person}{Girish Sastry}, \bibinfo{person}{Amanda Askell}, {et~al\mbox{.}}} \bibinfo{year}{2020}\natexlab{}.
\newblock \showarticletitle{{Language Models are Few-Shot Learners}}.
\newblock \bibinfo{journal}{\emph{Proc. of NeurIPS}}  \bibinfo{volume}{33} (\bibinfo{year}{2020}), \bibinfo{pages}{1877--1901}.
\newblock


\bibitem[Cai et~al\mbox{.}(2024)]%
        {cai2024self}
\bibfield{author}{\bibinfo{person}{Kaiwen Cai}, \bibinfo{person}{Zhekai Duan}, \bibinfo{person}{Gaowen Liu}, \bibinfo{person}{Charles Fleming}, {and} \bibinfo{person}{Chris~Xiaoxuan Lu}.} \bibinfo{year}{2024}\natexlab{}.
\newblock \showarticletitle{{Self-Adapting Large Visual-Language Models to Edge Devices across Visual Modalities}}.
\newblock \bibinfo{journal}{\emph{arXiv preprint arXiv:2403.04908}} (\bibinfo{year}{2024}).
\newblock


\bibitem[Cao et~al\mbox{.}(2022)]%
        {cao2022cross}
\bibfield{author}{\bibinfo{person}{Dongjiang Cao}, \bibinfo{person}{Ruofeng Liu}, \bibinfo{person}{Hao Li}, \bibinfo{person}{Shuai Wang}, \bibinfo{person}{Wenchao Jiang}, {and} \bibinfo{person}{Chris~Xiaoxuan Lu}.} \bibinfo{year}{2022}\natexlab{}.
\newblock \showarticletitle{{Cross Vision-RF Gait Re-identification with Low-cost RGB-D Cameras and mmWave Radars}}.
\newblock \bibinfo{journal}{\emph{Proc. of ACM UbiComp}} \bibinfo{volume}{6}, \bibinfo{number}{3} (\bibinfo{year}{2022}), \bibinfo{pages}{1--25}.
\newblock


\bibitem[Chakraborty et~al\mbox{.}(2022)]%
        {chakraborty2022diat}
\bibfield{author}{\bibinfo{person}{Mainak Chakraborty}, \bibinfo{person}{Harish~C Kumawat}, \bibinfo{person}{Sunita~Vikrant Dhavale}, {et~al\mbox{.}}} \bibinfo{year}{2022}\natexlab{}.
\newblock \showarticletitle{{DIAT-RadHARNet: A Lightweight DCNN for Radar based Classification of Human Suspicious Activities}}.
\newblock \bibinfo{journal}{\emph{IEEE Transactions on Instrumentation and Measurement}}  \bibinfo{volume}{71} (\bibinfo{year}{2022}), \bibinfo{pages}{1--10}.
\newblock


\bibitem[Chen et~al\mbox{.}(2021b)]%
        {chen2021magx}
\bibfield{author}{\bibinfo{person}{Dongyao Chen}, \bibinfo{person}{Mingke Wang}, \bibinfo{person}{Chenxi He}, \bibinfo{person}{Qing Luo}, \bibinfo{person}{Yasha Iravantchi}, \bibinfo{person}{Alanson Sample}, \bibinfo{person}{Kang~G Shin}, {and} \bibinfo{person}{Xinbing Wang}.} \bibinfo{year}{2021}\natexlab{b}.
\newblock \showarticletitle{{MagX: Wearable, Untethered Hands Tracking with Passive Magnets}}. In \bibinfo{booktitle}{\emph{Proc. of the 27th ACM MobiCom}}. \bibinfo{pages}{269--282}.
\newblock


\bibitem[Chen et~al\mbox{.}(2020)]%
        {chen2020simple}
\bibfield{author}{\bibinfo{person}{Ting Chen}, \bibinfo{person}{Simon Kornblith}, \bibinfo{person}{Mohammad Norouzi}, {and} \bibinfo{person}{Geoffrey Hinton}.} \bibinfo{year}{2020}\natexlab{}.
\newblock \showarticletitle{{A Simple Framework for Contrastive Learning of Visual Representations}}. In \bibinfo{booktitle}{\emph{Proc. of ICML}}. PMLR, \bibinfo{pages}{1597--1607}.
\newblock


\bibitem[Chen et~al\mbox{.}(2021a)]%
        {chen2021rf}
\bibfield{author}{\bibinfo{person}{Zhe Chen}, \bibinfo{person}{Chao Cai}, \bibinfo{person}{Tianyue Zheng}, \bibinfo{person}{Jun Luo}, \bibinfo{person}{Jie Xiong}, {and} \bibinfo{person}{Xin Wang}.} \bibinfo{year}{2021}\natexlab{a}.
\newblock \showarticletitle{{RF-based Human Activity Recognition using Signal Adapted Convolutional Neural Network}}.
\newblock \bibinfo{journal}{\emph{IEEE Transactions on Mobile Computing}} \bibinfo{volume}{22}, \bibinfo{number}{1} (\bibinfo{year}{2021}), \bibinfo{pages}{487--499}.
\newblock


\bibitem[Chen et~al\mbox{.}(2023)]%
        {chen2023wider}
\bibfield{author}{\bibinfo{person}{Zhe Chen}, \bibinfo{person}{Tianyue Zheng}, \bibinfo{person}{Chao Cai}, \bibinfo{person}{Yue Gao}, \bibinfo{person}{Pengfei Hu}, {and} \bibinfo{person}{Jun Luo}.} \bibinfo{year}{2023}\natexlab{}.
\newblock \showarticletitle{{Wider is Better? Contact-free Vibration Sensing via Different COTS-RF Technologies}}. In \bibinfo{booktitle}{\emph{Proc. of IEEE INFOCOM}}. IEEE, \bibinfo{pages}{1--10}.
\newblock


\bibitem[Chen et~al\mbox{.}(2021d)]%
        {chen2021movi}
\bibfield{author}{\bibinfo{person}{Zhe Chen}, \bibinfo{person}{Tianyue Zheng}, \bibinfo{person}{Chao Cai}, {and} \bibinfo{person}{Jun Luo}.} \bibinfo{year}{2021}\natexlab{d}.
\newblock \showarticletitle{{MoVi-Fi: Motion-robust Vital Signs Waveform Recovery via Deep Interpreted RF Sensing}}. In \bibinfo{booktitle}{\emph{Proc. of the 27th ACM MobiCom}}. \bibinfo{pages}{392--405}.
\newblock


\bibitem[Chen et~al\mbox{.}(2021c)]%
        {chen2021octopus}
\bibfield{author}{\bibinfo{person}{Zhe Chen}, \bibinfo{person}{Tianyue Zheng}, {and} \bibinfo{person}{Jun Luo}.} \bibinfo{year}{2021}\natexlab{c}.
\newblock \showarticletitle{{Octopus: a practical and versatile wideband MIMO sensing platform}}. In \bibinfo{booktitle}{\emph{Proc. of the 27th ACM MobiCom}}. \bibinfo{pages}{601--614}.
\newblock


\bibitem[Chi et~al\mbox{.}(2018)]%
        {chi2018ear}
\bibfield{author}{\bibinfo{person}{Zicheng Chi}, \bibinfo{person}{Yao Yao}, \bibinfo{person}{Tiantian Xie}, \bibinfo{person}{Xin Liu}, \bibinfo{person}{Zhichuan Huang}, \bibinfo{person}{Wei Wang}, {and} \bibinfo{person}{Ting Zhu}.} \bibinfo{year}{2018}\natexlab{}.
\newblock \showarticletitle{{EAR: Exploiting Uncontrollable Ambient RF Signals in Heterogeneous Networks for Gesture Recognition}}. In \bibinfo{booktitle}{\emph{Proc. of the 16th ACM SenSys}}. \bibinfo{pages}{237--249}.
\newblock


\bibitem[Dang et~al\mbox{.}(2020)]%
        {dang2020sensor}
\bibfield{author}{\bibinfo{person}{L~Minh Dang}, \bibinfo{person}{Kyungbok Min}, \bibinfo{person}{Hanxiang Wang}, \bibinfo{person}{Md~Jalil Piran}, \bibinfo{person}{Cheol~Hee Lee}, {and} \bibinfo{person}{Hyeonjoon Moon}.} \bibinfo{year}{2020}\natexlab{}.
\newblock \showarticletitle{{Sensor-Based and Vision-Based Human Activity Recognition: A Comprehensive Survey}}.
\newblock \bibinfo{journal}{\emph{Pattern Recognition}}  \bibinfo{volume}{108} (\bibinfo{year}{2020}), \bibinfo{pages}{107561}.
\newblock


\bibitem[Ding et~al\mbox{.}(2020)]%
        {ding2020rf}
\bibfield{author}{\bibinfo{person}{Shuya Ding}, \bibinfo{person}{Zhe Chen}, \bibinfo{person}{Tianyue Zheng}, {and} \bibinfo{person}{Jun Luo}.} \bibinfo{year}{2020}\natexlab{}.
\newblock \showarticletitle{{RF-Net: A Unified Meta-Learning Framework for RF-enabled One-Shot Human Activity Recognition}}. In \bibinfo{booktitle}{\emph{Proc. of the 18th ACM SenSys}}. \bibinfo{pages}{517--530}.
\newblock


\bibitem[Dosovitskiy et~al\mbox{.}(2020)]%
        {dosovitskiy2020image}
\bibfield{author}{\bibinfo{person}{Alexey Dosovitskiy}, \bibinfo{person}{Lucas Beyer}, \bibinfo{person}{Alexander Kolesnikov}, \bibinfo{person}{Dirk Weissenborn}, \bibinfo{person}{Xiaohua Zhai}, \bibinfo{person}{Thomas Unterthiner}, \bibinfo{person}{Mostafa Dehghani}, \bibinfo{person}{Matthias Minderer}, \bibinfo{person}{Georg Heigold}, \bibinfo{person}{Sylvain Gelly}, {et~al\mbox{.}}} \bibinfo{year}{2020}\natexlab{}.
\newblock \showarticletitle{{An Image Is Worth 16x16 Words: Transformers for Image Recognition at Scale}}.
\newblock \bibinfo{journal}{\emph{arXiv preprint arXiv:2010.11929}} (\bibinfo{year}{2020}).
\newblock


\bibitem[Esmaeilpour et~al\mbox{.}(2022)]%
        {esmaeilpour2022zero}
\bibfield{author}{\bibinfo{person}{Sepideh Esmaeilpour}, \bibinfo{person}{Bing Liu}, \bibinfo{person}{Eric Robertson}, {and} \bibinfo{person}{Lei Shu}.} \bibinfo{year}{2022}\natexlab{}.
\newblock \showarticletitle{{Zero-Shot Out-of-Distribution Detection Based on the Pre-trained Model CLIP}}. In \bibinfo{booktitle}{\emph{Proc. of AAAI}}, Vol.~\bibinfo{volume}{36}. \bibinfo{pages}{6568--6576}.
\newblock


\bibitem[Fan et~al\mbox{.}(2020)]%
        {fan2020home}
\bibfield{author}{\bibinfo{person}{Lijie Fan}, \bibinfo{person}{Tianhong Li}, \bibinfo{person}{Yuan Yuan}, {and} \bibinfo{person}{Dina Katabi}.} \bibinfo{year}{2020}\natexlab{}.
\newblock \showarticletitle{{In-Home Daily-Life Captioning Using Radio Signals}}. In \bibinfo{booktitle}{\emph{Proc. of the 16th ECCV}}. Springer, \bibinfo{pages}{105--123}.
\newblock


\bibitem[Fang et~al\mbox{.}(2023)]%
        {fang2023eva}
\bibfield{author}{\bibinfo{person}{Yuxin Fang}, \bibinfo{person}{Wen Wang}, \bibinfo{person}{Binhui Xie}, \bibinfo{person}{Quan Sun}, \bibinfo{person}{Ledell Wu}, \bibinfo{person}{Xinggang Wang}, \bibinfo{person}{Tiejun Huang}, \bibinfo{person}{Xinlong Wang}, {and} \bibinfo{person}{Yue Cao}.} \bibinfo{year}{2023}\natexlab{}.
\newblock \showarticletitle{{EVA: Exploring the Limits of Masked Visual Representation Learning at Scale}}. In \bibinfo{booktitle}{\emph{Proc. of IEEE/CVF CVPR}}. \bibinfo{pages}{19358--19369}.
\newblock


\bibitem[Ferlini et~al\mbox{.}(2021)]%
        {ferlini2021eargate}
\bibfield{author}{\bibinfo{person}{Andrea Ferlini}, \bibinfo{person}{Dong Ma}, \bibinfo{person}{Robert Harle}, {and} \bibinfo{person}{Cecilia Mascolo}.} \bibinfo{year}{2021}\natexlab{}.
\newblock \showarticletitle{{EarGate: Gait-based User Identification with In-ear Microphones}}. In \bibinfo{booktitle}{\emph{Proc. of the 27th ACM MobiCom}}. \bibinfo{pages}{337--349}.
\newblock


\bibitem[Ganin and Lempitsky(2015)]%
        {ganin2015unsupervised}
\bibfield{author}{\bibinfo{person}{Yaroslav Ganin} {and} \bibinfo{person}{Victor Lempitsky}.} \bibinfo{year}{2015}\natexlab{}.
\newblock \showarticletitle{{Unsupervised Domain Adaptation by Backpropagation}}. In \bibinfo{booktitle}{\emph{Proc. of ICML}}. PMLR, \bibinfo{pages}{1180--1189}.
\newblock


\bibitem[Ganin et~al\mbox{.}(2016)]%
        {ganin2016domain}
\bibfield{author}{\bibinfo{person}{Yaroslav Ganin}, \bibinfo{person}{Evgeniya Ustinova}, \bibinfo{person}{Hana Ajakan}, \bibinfo{person}{Pascal Germain}, \bibinfo{person}{Hugo Larochelle}, \bibinfo{person}{Fran{\c{c}}ois Laviolette}, \bibinfo{person}{Mario March}, {and} \bibinfo{person}{Victor Lempitsky}.} \bibinfo{year}{2016}\natexlab{}.
\newblock \showarticletitle{{Domain-adversarial Training of Neural Networks}}.
\newblock \bibinfo{journal}{\emph{Journal of Machine Learning Research}} \bibinfo{volume}{17}, \bibinfo{number}{59} (\bibinfo{year}{2016}), \bibinfo{pages}{1--35}.
\newblock


\bibitem[Gao et~al\mbox{.}(2022)]%
        {gao2022towards}
\bibfield{author}{\bibinfo{person}{Ruiyang Gao}, \bibinfo{person}{Wenwei Li}, \bibinfo{person}{Yaxiong Xie}, \bibinfo{person}{Enze Yi}, \bibinfo{person}{Leye Wang}, \bibinfo{person}{Dan Wu}, {and} \bibinfo{person}{Daqing Zhang}.} \bibinfo{year}{2022}\natexlab{}.
\newblock \showarticletitle{{Towards Robust Gesture Recognition by Characterizing the Sensing Quality of WiFi Signals}}.
\newblock \bibinfo{journal}{\emph{Proc. of ACM UbiComp}} \bibinfo{volume}{6}, \bibinfo{number}{1} (\bibinfo{year}{2022}), \bibinfo{pages}{1--26}.
\newblock


\bibitem[Gong et~al\mbox{.}(2019)]%
        {gong2019metasense}
\bibfield{author}{\bibinfo{person}{Taesik Gong}, \bibinfo{person}{Yeonsu Kim}, \bibinfo{person}{Jinwoo Shin}, {and} \bibinfo{person}{Sung-Ju Lee}.} \bibinfo{year}{2019}\natexlab{}.
\newblock \showarticletitle{{MetaSense: Few-Shot Adaptation to Untrained Conditions in Deep Mobile Sensing}}. In \bibinfo{booktitle}{\emph{Proc. of the 17th ACM SenSys}}. \bibinfo{pages}{110--123}.
\newblock


\bibitem[Hao et~al\mbox{.}(2018)]%
        {hao2018recognizing}
\bibfield{author}{\bibinfo{person}{Jianguo Hao}, \bibinfo{person}{Abdenour Bouzouane}, {and} \bibinfo{person}{S{\'e}bastien Gaboury}.} \bibinfo{year}{2018}\natexlab{}.
\newblock \showarticletitle{{Recognizing Multi-Resident Activities in Non-Intrusive Sensor-Based Smart Homes by Formal Concept Analysis}}.
\newblock \bibinfo{journal}{\emph{Neurocomputing}}  \bibinfo{volume}{318} (\bibinfo{year}{2018}), \bibinfo{pages}{75--89}.
\newblock


\bibitem[He et~al\mbox{.}(2020)]%
        {he2020momentum}
\bibfield{author}{\bibinfo{person}{Kaiming He}, \bibinfo{person}{Haoqi Fan}, \bibinfo{person}{Yuxin Wu}, \bibinfo{person}{Saining Xie}, {and} \bibinfo{person}{Ross Girshick}.} \bibinfo{year}{2020}\natexlab{}.
\newblock \showarticletitle{{Momentum Contrast for Unsupervised Visual Representation Learning}}. In \bibinfo{booktitle}{\emph{Proc. of IEEE/CVF CVPR}}. \bibinfo{pages}{9729--9738}.
\newblock


\bibitem[Hinton et~al\mbox{.}(2015)]%
        {hinton2015distilling}
\bibfield{author}{\bibinfo{person}{Geoffrey Hinton}, \bibinfo{person}{Oriol Vinyals}, {and} \bibinfo{person}{Jeff Dean}.} \bibinfo{year}{2015}\natexlab{}.
\newblock \showarticletitle{{Distilling the Knowledge in a Neural Network}}.
\newblock \bibinfo{journal}{\emph{arXiv preprint arXiv:1503.02531}} (\bibinfo{year}{2015}).
\newblock


\bibitem[Hu et~al\mbox{.}(2024)]%
        {hu2024m}
\bibfield{author}{\bibinfo{person}{Jingyang Hu}, \bibinfo{person}{Hongbo Jiang}, \bibinfo{person}{Tianyue Zheng}, \bibinfo{person}{Jingzhi Hu}, \bibinfo{person}{Hongbo Wang}, \bibinfo{person}{Hangcheng Cao}, \bibinfo{person}{Zhe Chen}, {and} \bibinfo{person}{Jun Luo}.} \bibinfo{year}{2024}\natexlab{}.
\newblock \showarticletitle{M2-Fi: Multi-person Respiration Monitoring via Handheld WiFi Devices}.
\newblock \bibinfo{journal}{\emph{Proc. of IEEE INFOCOM}} (\bibinfo{year}{2024}).
\newblock


\bibitem[Hu et~al\mbox{.}(2023)]%
        {hu2023muse}
\bibfield{author}{\bibinfo{person}{Jingzhi Hu}, \bibinfo{person}{Tianyue Zheng}, \bibinfo{person}{Zhe Chen}, \bibinfo{person}{Hongbo Wang}, {and} \bibinfo{person}{Jun Luo}.} \bibinfo{year}{2023}\natexlab{}.
\newblock \showarticletitle{{MUSE-Fi: Contactless MUti-person SEnsing Exploiting Near-field Wi-Fi Channel Variation}}. In \bibinfo{booktitle}{\emph{Proc. of the 29th ACM MobiCom}}. \bibinfo{pages}{1--15}.
\newblock


\bibitem[Huang et~al\mbox{.}(2024)]%
        {10339891}
\bibfield{author}{\bibinfo{person}{Jinyang Huang}, \bibinfo{person}{Bin Liu}, \bibinfo{person}{Chenglin Miao}, \bibinfo{person}{Xiang Zhang}, \bibinfo{person}{Jianchun Liu}, \bibinfo{person}{Lu Su}, \bibinfo{person}{Zhi Liu}, {and} \bibinfo{person}{Yu Gu}.} \bibinfo{year}{2024}\natexlab{}.
\newblock \showarticletitle{{PhyFinAtt: An Undetectable Attack Framework Against PHY Layer Fingerprint-Based WiFi Authentication}}.
\newblock \bibinfo{journal}{\emph{IEEE Transactions on Mobile Computing}} \bibinfo{volume}{23}, \bibinfo{number}{7} (\bibinfo{year}{2024}), \bibinfo{pages}{7753--7770}.
\newblock


\bibitem[Jiang et~al\mbox{.}(2018)]%
        {jiang2018towards}
\bibfield{author}{\bibinfo{person}{Wenjun Jiang}, \bibinfo{person}{Chenglin Miao}, \bibinfo{person}{Fenglong Ma}, \bibinfo{person}{Shuochao Yao}, \bibinfo{person}{Yaqing Wang}, \bibinfo{person}{Ye Yuan}, \bibinfo{person}{Hongfei Xue}, \bibinfo{person}{Chen Song}, \bibinfo{person}{Xin Ma}, \bibinfo{person}{Dimitrios Koutsonikolas}, {et~al\mbox{.}}} \bibinfo{year}{2018}\natexlab{}.
\newblock \showarticletitle{{Towards Environment Independent Device Free Human Activity Recognition}}. In \bibinfo{booktitle}{\emph{Proc. of the 24th ACM MobiCom}}. \bibinfo{pages}{289--304}.
\newblock


\bibitem[Kirillov et~al\mbox{.}(2023)]%
        {kirillov2023segment}
\bibfield{author}{\bibinfo{person}{Alexander Kirillov}, \bibinfo{person}{Eric Mintun}, \bibinfo{person}{Nikhila Ravi}, \bibinfo{person}{Hanzi Mao}, \bibinfo{person}{Chloe Rolland}, \bibinfo{person}{Laura Gustafson}, \bibinfo{person}{Tete Xiao}, \bibinfo{person}{Spencer Whitehead}, \bibinfo{person}{Alexander~C Berg}, \bibinfo{person}{Wan-Yen Lo}, {et~al\mbox{.}}} \bibinfo{year}{2023}\natexlab{}.
\newblock \showarticletitle{{Segment Anything}}.
\newblock \bibinfo{journal}{\emph{arXiv preprint arXiv:2304.02643}} (\bibinfo{year}{2023}).
\newblock


\bibitem[Li et~al\mbox{.}(2023)]%
        {li2023human}
\bibfield{author}{\bibinfo{person}{Xiaoxiong Li}, \bibinfo{person}{Si Chen}, \bibinfo{person}{Shuning Zhang}, \bibinfo{person}{Linsheng Hou}, \bibinfo{person}{Yuying Zhu}, {and} \bibinfo{person}{Zelong Xiao}.} \bibinfo{year}{2023}\natexlab{}.
\newblock \showarticletitle{{Human Activity Recognition Using IR-UWB Radar: A Lightweight Transformer Approach}}.
\newblock \bibinfo{journal}{\emph{IEEE Geoscience and Remote Sensing Letters}} (\bibinfo{year}{2023}).
\newblock


\bibitem[Li et~al\mbox{.}(2021)]%
        {li2021semisupervised}
\bibfield{author}{\bibinfo{person}{Xinyu Li}, \bibinfo{person}{Yuan He}, \bibinfo{person}{Francesco Fioranelli}, {and} \bibinfo{person}{Xiaojun Jing}.} \bibinfo{year}{2021}\natexlab{}.
\newblock \showarticletitle{{Semisupervised Human Activity Recognition with Radar Micro-Doppler Signatures}}.
\newblock \bibinfo{journal}{\emph{IEEE Transactions on Geoscience and Remote Sensing}}  \bibinfo{volume}{60} (\bibinfo{year}{2021}), \bibinfo{pages}{1--12}.
\newblock


\bibitem[Liu et~al\mbox{.}(2019)]%
        {liu2019real}
\bibfield{author}{\bibinfo{person}{Yang Liu}, \bibinfo{person}{Zhenjiang Li}, \bibinfo{person}{Zhidan Liu}, {and} \bibinfo{person}{Kaishun Wu}.} \bibinfo{year}{2019}\natexlab{}.
\newblock \showarticletitle{{Real-time Arm Skeleton Tracking and Gesture Inference Tolerant to Missing Wearable Sensors}}. In \bibinfo{booktitle}{\emph{Proc. of the 17th ACM MobiSys}}. \bibinfo{pages}{287--299}.
\newblock


\bibitem[Melgarejo et~al\mbox{.}(2014)]%
        {melgarejo2014leveraging}
\bibfield{author}{\bibinfo{person}{Pedro Melgarejo}, \bibinfo{person}{Xinyu Zhang}, \bibinfo{person}{Parameswaran Ramanathan}, {and} \bibinfo{person}{David Chu}.} \bibinfo{year}{2014}\natexlab{}.
\newblock \showarticletitle{{Leveraging Directional Antenna Capabilities for Fine-grained Gesture Recognition}}. In \bibinfo{booktitle}{\emph{Proc. of ACM UbiComp}}. \bibinfo{pages}{541--551}.
\newblock


\bibitem[Meneghello et~al\mbox{.}(2022)]%
        {meneghello2022sharp}
\bibfield{author}{\bibinfo{person}{Francesca Meneghello}, \bibinfo{person}{Domenico Garlisi}, \bibinfo{person}{Nicol{\`o} Dal~Fabbro}, \bibinfo{person}{Ilenia Tinnirello}, {and} \bibinfo{person}{Michele Rossi}.} \bibinfo{year}{2022}\natexlab{}.
\newblock \showarticletitle{{ShARP: Environment and Person Independent Activity Recognition with Commodity IEEE 802.11 Access Points}}.
\newblock \bibinfo{journal}{\emph{IEEE Transactions on Mobile Computing}} (\bibinfo{year}{2022}).
\newblock


\bibitem[{Microsoft}(2020)]%
        {microsoft_kinect}
\bibfield{author}{\bibinfo{person}{{Microsoft}}.} \bibinfo{year}{2020}\natexlab{}.
\newblock \bibinfo{title}{{Kinect Sensor}}.
\newblock \bibinfo{howpublished}{\url{https://developer.microsoft.com/en-us/windows/kinect/}}.
\newblock
\newblock
\shownote{Accessed: 2020-09-29}.


\bibitem[Minderer et~al\mbox{.}(2022)]%
        {minderer2022simple}
\bibfield{author}{\bibinfo{person}{Matthias Minderer}, \bibinfo{person}{Alexey Gritsenko}, \bibinfo{person}{Austin Stone}, \bibinfo{person}{Maxim Neumann}, \bibinfo{person}{Dirk Weissenborn}, \bibinfo{person}{Alexey Dosovitskiy}, \bibinfo{person}{Aravindh Mahendran}, \bibinfo{person}{Anurag Arnab}, \bibinfo{person}{Mostafa Dehghani}, \bibinfo{person}{Zhuoran Shen}, {et~al\mbox{.}}} \bibinfo{year}{2022}\natexlab{}.
\newblock \showarticletitle{{Simple Open-Vocabulary Object Detection with Vision Transformers}}. In \bibinfo{booktitle}{\emph{Proc. of ECCV}}. Springer, \bibinfo{pages}{728--755}.
\newblock


\bibitem[Niu et~al\mbox{.}(2018)]%
        {niu2018boosting}
\bibfield{author}{\bibinfo{person}{Kai Niu}, \bibinfo{person}{Fusang Zhang}, \bibinfo{person}{Jie Xiong}, \bibinfo{person}{Xiang Li}, \bibinfo{person}{Enze Yi}, {and} \bibinfo{person}{Daqing Zhang}.} \bibinfo{year}{2018}\natexlab{}.
\newblock \showarticletitle{{Boosting Fine-grained Activity Sensing by Embracing Wireless Multipath Effects}}. In \bibinfo{booktitle}{\emph{Proc. of the 14th ACM CoNEXT}}. \bibinfo{pages}{139--151}.
\newblock


\bibitem[Ouyang et~al\mbox{.}(2021)]%
        {ouyang2021clusterfl}
\bibfield{author}{\bibinfo{person}{Xiaomin Ouyang}, \bibinfo{person}{Zhiyuan Xie}, \bibinfo{person}{Jiayu Zhou}, \bibinfo{person}{Jianwei Huang}, {and} \bibinfo{person}{Guoliang Xing}.} \bibinfo{year}{2021}\natexlab{}.
\newblock \showarticletitle{{ClusterFL: A Similarity-aware Federated Learning System for Human Activity Recognition}}. In \bibinfo{booktitle}{\emph{Proc. of the 19th ACM MobiSys}}. \bibinfo{pages}{54--66}.
\newblock


\bibitem[Palipana et~al\mbox{.}(2021)]%
        {palipana2021pantomime}
\bibfield{author}{\bibinfo{person}{Sameera Palipana}, \bibinfo{person}{Dariush Salami}, \bibinfo{person}{Luis~A Leiva}, {and} \bibinfo{person}{Stephan Sigg}.} \bibinfo{year}{2021}\natexlab{}.
\newblock \showarticletitle{{Pantomime: Mid-Air Gesture Recognition with Sparse Millimeter-Wave Radar Point Clouds}}.
\newblock \bibinfo{journal}{\emph{Proc. of ACM UbiComp}} \bibinfo{volume}{5}, \bibinfo{number}{1} (\bibinfo{year}{2021}), \bibinfo{pages}{1--27}.
\newblock


\bibitem[Park et~al\mbox{.}(2009)]%
        {park2009cadence}
\bibfield{author}{\bibinfo{person}{Hyung~O Park}, \bibinfo{person}{Alireza~A Dibazar}, {and} \bibinfo{person}{Theodore~W Berger}.} \bibinfo{year}{2009}\natexlab{}.
\newblock \showarticletitle{{Cadence Analysis of Temporal Gait Patterns for Seismic Discrimination Between Human and Quadruped Footsteps}}. In \bibinfo{booktitle}{\emph{Proc. of IEEE ICASSP}}. IEEE, \bibinfo{pages}{1749--1752}.
\newblock


\bibitem[Peng et~al\mbox{.}(2019)]%
        {peng2019correlation}
\bibfield{author}{\bibinfo{person}{Baoyun Peng}, \bibinfo{person}{Xiao Jin}, \bibinfo{person}{Jiaheng Liu}, \bibinfo{person}{Dongsheng Li}, \bibinfo{person}{Yichao Wu}, \bibinfo{person}{Yu Liu}, \bibinfo{person}{Shunfeng Zhou}, {and} \bibinfo{person}{Zhaoning Zhang}.} \bibinfo{year}{2019}\natexlab{}.
\newblock \showarticletitle{{Correlation Congruence for Knowledge Distillation}}. In \bibinfo{booktitle}{\emph{Proc. of IEEE/CVF ICCV}}. \bibinfo{pages}{5007--5016}.
\newblock


\bibitem[Qi et~al\mbox{.}(2017a)]%
        {qi2017pointnet}
\bibfield{author}{\bibinfo{person}{Charles~R Qi}, \bibinfo{person}{Hao Su}, \bibinfo{person}{Kaichun Mo}, {and} \bibinfo{person}{Leonidas~J Guibas}.} \bibinfo{year}{2017}\natexlab{a}.
\newblock \showarticletitle{{PointNet: Deep Learning on Point Sets for 3D Classification and Segmentation}}. In \bibinfo{booktitle}{\emph{Proc. of IEEE/CVF CVPR}}. \bibinfo{pages}{652--660}.
\newblock


\bibitem[Qi et~al\mbox{.}(2017b)]%
        {qi2017pointnet++}
\bibfield{author}{\bibinfo{person}{Charles~Ruizhongtai Qi}, \bibinfo{person}{Li Yi}, \bibinfo{person}{Hao Su}, {and} \bibinfo{person}{Leonidas~J Guibas}.} \bibinfo{year}{2017}\natexlab{b}.
\newblock \showarticletitle{{PointNet++: Deep Hierarchical Feature Learning on Point Sets in a Metric Space}}.
\newblock \bibinfo{journal}{\emph{Proc. of NeurIPS}}  \bibinfo{volume}{30} (\bibinfo{year}{2017}).
\newblock


\bibitem[Radford et~al\mbox{.}(2021)]%
        {radford2021learning}
\bibfield{author}{\bibinfo{person}{Alec Radford}, \bibinfo{person}{Jong~Wook Kim}, \bibinfo{person}{Chris Hallacy}, \bibinfo{person}{Aditya Ramesh}, \bibinfo{person}{Gabriel Goh}, \bibinfo{person}{Sandhini Agarwal}, \bibinfo{person}{Girish Sastry}, \bibinfo{person}{Amanda Askell}, \bibinfo{person}{Pamela Mishkin}, \bibinfo{person}{Jack Clark}, {et~al\mbox{.}}} \bibinfo{year}{2021}\natexlab{}.
\newblock \showarticletitle{{Learning Transferable Visual Models from Natural Language Supervision}}. In \bibinfo{booktitle}{\emph{Proc. of ICML}}. PMLR, \bibinfo{pages}{8748--8763}.
\newblock


\bibitem[Ramesh et~al\mbox{.}(2022)]%
        {ramesh2022hierarchical}
\bibfield{author}{\bibinfo{person}{Aditya Ramesh}, \bibinfo{person}{Prafulla Dhariwal}, \bibinfo{person}{Alex Nichol}, \bibinfo{person}{Casey Chu}, {and} \bibinfo{person}{Mark Chen}.} \bibinfo{year}{2022}\natexlab{}.
\newblock \showarticletitle{{Hierarchical Text-Conditional Image Generation with CLIP Latents}}.
\newblock \bibinfo{journal}{\emph{arXiv preprint arXiv:2204.06125}} \bibinfo{volume}{1}, \bibinfo{number}{2} (\bibinfo{year}{2022}), \bibinfo{pages}{3}.
\newblock


\bibitem[Ramesh et~al\mbox{.}(2021)]%
        {ramesh2021zero}
\bibfield{author}{\bibinfo{person}{Aditya Ramesh}, \bibinfo{person}{Mikhail Pavlov}, \bibinfo{person}{Gabriel Goh}, \bibinfo{person}{Scott Gray}, \bibinfo{person}{Chelsea Voss}, \bibinfo{person}{Alec Radford}, \bibinfo{person}{Mark Chen}, {and} \bibinfo{person}{Ilya Sutskever}.} \bibinfo{year}{2021}\natexlab{}.
\newblock \showarticletitle{{Zero-Shot Text-to-Image Generation}}. In \bibinfo{booktitle}{\emph{Proc. of ICML}}. PMLR, \bibinfo{pages}{8821--8831}.
\newblock


\bibitem[Ren et~al\mbox{.}(2023)]%
        {ren2022delving}
\bibfield{author}{\bibinfo{person}{Shuhuai Ren}, \bibinfo{person}{Lei Li}, \bibinfo{person}{Xuancheng Ren}, \bibinfo{person}{Guangxiang Zhao}, {and} \bibinfo{person}{Xu Sun}.} \bibinfo{year}{2023}\natexlab{}.
\newblock \showarticletitle{{Delving into the Openness of CLIP}}. In \bibinfo{booktitle}{\emph{Proc. of ACL}}.
\newblock


\bibitem[Rodomagoulakis et~al\mbox{.}(2016)]%
        {rodomagoulakis2016multimodal}
\bibfield{author}{\bibinfo{person}{Isidoros Rodomagoulakis}, \bibinfo{person}{Nikolaos Kardaris}, \bibinfo{person}{Vassilis Pitsikalis}, \bibinfo{person}{E Mavroudi}, \bibinfo{person}{Athanasios Katsamanis}, \bibinfo{person}{Antigoni Tsiami}, {and} \bibinfo{person}{Petros Maragos}.} \bibinfo{year}{2016}\natexlab{}.
\newblock \showarticletitle{{Multimodal Human Action Recognition in Assistive Human-Robot Interaction}}. In \bibinfo{booktitle}{\emph{Proc. of IEEE ICASSP}}. IEEE, \bibinfo{pages}{2702--2706}.
\newblock


\bibitem[Salami et~al\mbox{.}(2022)]%
        {salami2022tesla}
\bibfield{author}{\bibinfo{person}{Dariush Salami}, \bibinfo{person}{Ramin Hasibi}, \bibinfo{person}{Sameera Palipana}, \bibinfo{person}{Petar Popovski}, \bibinfo{person}{Tom Michoel}, {and} \bibinfo{person}{Stephan Sigg}.} \bibinfo{year}{2022}\natexlab{}.
\newblock \showarticletitle{{Tesla-Rapture: A Lightweight Gesture Recognition System from mmWave Radar Sparse Point Clouds}}.
\newblock \bibinfo{journal}{\emph{IEEE Transactions on Mobile Computing}} (\bibinfo{year}{2022}).
\newblock


\bibitem[Seifert et~al\mbox{.}(2019)]%
        {seifert2019toward}
\bibfield{author}{\bibinfo{person}{Ann-Kathrin Seifert}, \bibinfo{person}{Moeness~G Amin}, {and} \bibinfo{person}{Abdelhak~M Zoubir}.} \bibinfo{year}{2019}\natexlab{}.
\newblock \showarticletitle{{Toward Unobtrusive In-home Gait Analysis Based on Radar Micro-Doppler Signatures}}.
\newblock \bibinfo{journal}{\emph{IEEE Transactions on Biomedical Engineering}} \bibinfo{volume}{66}, \bibinfo{number}{9} (\bibinfo{year}{2019}), \bibinfo{pages}{2629--2640}.
\newblock


\bibitem[Shtedritski et~al\mbox{.}(2023)]%
        {shtedritski2023does}
\bibfield{author}{\bibinfo{person}{Aleksandar Shtedritski}, \bibinfo{person}{Christian Rupprecht}, {and} \bibinfo{person}{Andrea Vedaldi}.} \bibinfo{year}{2023}\natexlab{}.
\newblock \showarticletitle{{What Does CLIP Know About a Red Circle? Visual Prompt Engineering for VLMs}}. In \bibinfo{booktitle}{\emph{Proc. of IEEE/CVF ICCV}}. \bibinfo{pages}{11987--11997}.
\newblock


\bibitem[Simonyan et~al\mbox{.}(2014)]%
        {simonyan2014deep}
\bibfield{author}{\bibinfo{person}{K Simonyan}, \bibinfo{person}{A Vedaldi}, {and} \bibinfo{person}{A Zisserman}.} \bibinfo{year}{2014}\natexlab{}.
\newblock \showarticletitle{{Deep Inside Convolutional Networks: Visualising Image Classification Models and Saliency Maps}}. In \bibinfo{booktitle}{\emph{Proc. of ICLR}}.
\newblock


\bibitem[Singh et~al\mbox{.}(2023)]%
        {singh2023depth}
\bibfield{author}{\bibinfo{person}{Akash~Deep Singh}, \bibinfo{person}{Yunhao Ba}, \bibinfo{person}{Ankur Sarker}, \bibinfo{person}{Howard Zhang}, \bibinfo{person}{Achuta Kadambi}, \bibinfo{person}{Stefano Soatto}, \bibinfo{person}{Mani Srivastava}, {and} \bibinfo{person}{Alex Wong}.} \bibinfo{year}{2023}\natexlab{}.
\newblock \showarticletitle{{Depth Estimation From Camera Image and mmWave Radar Point Cloud}}. In \bibinfo{booktitle}{\emph{Proc. of IEEE/CVF CVPR}}. \bibinfo{pages}{9275--9285}.
\newblock


\bibitem[Sun et~al\mbox{.}(2020)]%
        {sun2020wearable}
\bibfield{author}{\bibinfo{person}{Minglong Sun}, \bibinfo{person}{Amanda Watson}, {and} \bibinfo{person}{Gang Zhou}.} \bibinfo{year}{2020}\natexlab{}.
\newblock \showarticletitle{{Wearable Computing of Freezing of Gait in Parkinson's Disease: A Survey}}.
\newblock \bibinfo{journal}{\emph{Smart Health}}  \bibinfo{volume}{18} (\bibinfo{year}{2020}), \bibinfo{pages}{100143}.
\newblock


\bibitem[{Texas Instruments}(2020)]%
        {ti_iwr1443boost}
\bibfield{author}{\bibinfo{person}{{Texas Instruments}}.} \bibinfo{year}{2020}\natexlab{}.
\newblock \bibinfo{title}{{IWR1443BOOST}}.
\newblock \bibinfo{howpublished}{\url{https://www.ti.com/tool/IWR1443BOOST}}.
\newblock
\newblock
\shownote{Accessed: 2020-09-29}.


\bibitem[Tian et~al\mbox{.}(2019)]%
        {tian2019contrastive}
\bibfield{author}{\bibinfo{person}{Yonglong Tian}, \bibinfo{person}{Dilip Krishnan}, {and} \bibinfo{person}{Phillip Isola}.} \bibinfo{year}{2019}\natexlab{}.
\newblock \showarticletitle{{Contrastive Representation Distillation}}. In \bibinfo{booktitle}{\emph{Proc. of ICLR}}.
\newblock


\bibitem[Truong et~al\mbox{.}(2018)]%
        {truong2018capband}
\bibfield{author}{\bibinfo{person}{Hoang Truong}, \bibinfo{person}{Shuo Zhang}, \bibinfo{person}{Ufuk Muncuk}, \bibinfo{person}{Phuc Nguyen}, \bibinfo{person}{Nam Bui}, \bibinfo{person}{Anh Nguyen}, \bibinfo{person}{Qin Lv}, \bibinfo{person}{Kaushik Chowdhury}, \bibinfo{person}{Thang Dinh}, {and} \bibinfo{person}{Tam Vu}.} \bibinfo{year}{2018}\natexlab{}.
\newblock \showarticletitle{{Capband: Battery-Free Successive Capacitance Sensing Wristband for Hand Gesture Recognition}}. In \bibinfo{booktitle}{\emph{Proc. of the 16th ACM SenSys}}. \bibinfo{pages}{54--67}.
\newblock


\bibitem[Vaswani et~al\mbox{.}(2017)]%
        {vaswani2017attention}
\bibfield{author}{\bibinfo{person}{Ashish Vaswani}, \bibinfo{person}{Noam Shazeer}, \bibinfo{person}{Niki Parmar}, \bibinfo{person}{Jakob Uszkoreit}, \bibinfo{person}{Llion Jones}, \bibinfo{person}{Aidan~N Gomez}, \bibinfo{person}{{\L}ukasz Kaiser}, {and} \bibinfo{person}{Illia Polosukhin}.} \bibinfo{year}{2017}\natexlab{}.
\newblock \showarticletitle{{Attention Is All You Need}}.
\newblock \bibinfo{journal}{\emph{Proc. of NeurIPS}}  \bibinfo{volume}{30} (\bibinfo{year}{2017}).
\newblock


\bibitem[Wan et~al\mbox{.}(2020)]%
        {wan2020deep}
\bibfield{author}{\bibinfo{person}{Shaohua Wan}, \bibinfo{person}{Lianyong Qi}, \bibinfo{person}{Xiaolong Xu}, \bibinfo{person}{Chao Tong}, {and} \bibinfo{person}{Zonghua Gu}.} \bibinfo{year}{2020}\natexlab{}.
\newblock \showarticletitle{{Deep Learning Models for Real-Time Human Activity Recognition with Smartphones}}.
\newblock \bibinfo{journal}{\emph{Mobile Networks and Applications}}  \bibinfo{volume}{25} (\bibinfo{year}{2020}), \bibinfo{pages}{743--755}.
\newblock


\bibitem[Wang et~al\mbox{.}(2019)]%
        {wang2019ev}
\bibfield{author}{\bibinfo{person}{Yanxiang Wang}, \bibinfo{person}{Bowen Du}, \bibinfo{person}{Yiran Shen}, \bibinfo{person}{Kai Wu}, \bibinfo{person}{Guangrong Zhao}, \bibinfo{person}{Jianguo Sun}, {and} \bibinfo{person}{Hongkai Wen}.} \bibinfo{year}{2019}\natexlab{}.
\newblock \showarticletitle{{EV-Gait: Event-based Robust Gait Recognition using Dynamic Vision Sensors}}. In \bibinfo{booktitle}{\emph{Proc. of IEEE/CVF CVPR}}. \bibinfo{pages}{6358--6367}.
\newblock


\bibitem[Wortsman et~al\mbox{.}(2022)]%
        {wortsman2022robust}
\bibfield{author}{\bibinfo{person}{Mitchell Wortsman}, \bibinfo{person}{Gabriel Ilharco}, \bibinfo{person}{Jong~Wook Kim}, \bibinfo{person}{Mike Li}, \bibinfo{person}{Simon Kornblith}, \bibinfo{person}{Rebecca Roelofs}, \bibinfo{person}{Raphael~Gontijo Lopes}, \bibinfo{person}{Hannaneh Hajishirzi}, \bibinfo{person}{Ali Farhadi}, \bibinfo{person}{Hongseok Namkoong}, {et~al\mbox{.}}} \bibinfo{year}{2022}\natexlab{}.
\newblock \showarticletitle{{Robust fine-tuning of zero-shot models}}. In \bibinfo{booktitle}{\emph{Proc. of IEEE/CVF CVPR}}. \bibinfo{pages}{7959--7971}.
\newblock


\bibitem[Xie et~al\mbox{.}(2022)]%
        {xie2022deepvs}
\bibfield{author}{\bibinfo{person}{Zongxing Xie}, \bibinfo{person}{Hanrui Wang}, \bibinfo{person}{Song Han}, \bibinfo{person}{Elinor Schoenfeld}, {and} \bibinfo{person}{Fan Ye}.} \bibinfo{year}{2022}\natexlab{}.
\newblock \showarticletitle{{DeepVS: A Deep Learning Approach for RF-based Vital Signs Sensing}}. In \bibinfo{booktitle}{\emph{Proc. of the 13rd ACM BCB}}. \bibinfo{pages}{1--5}.
\newblock


\bibitem[Zhang et~al\mbox{.}(2023a)]%
        {zhang2023pi}
\bibfield{author}{\bibinfo{person}{Bo Zhang}, \bibinfo{person}{Boyu Jiang}, \bibinfo{person}{Rong Zheng}, \bibinfo{person}{Xiaoping Zhang}, \bibinfo{person}{Jun Li}, {and} \bibinfo{person}{Qiang Xu}.} \bibinfo{year}{2023}\natexlab{a}.
\newblock \showarticletitle{{Pi-Vimo: Physiology-inspired Robust Vital Sign Monitoring using mmWave Radars}}.
\newblock \bibinfo{journal}{\emph{ACM Transactions on Internet of Things}} \bibinfo{volume}{4}, \bibinfo{number}{2} (\bibinfo{year}{2023}), \bibinfo{pages}{1--27}.
\newblock


\bibitem[Zhang et~al\mbox{.}(2023b)]%
        {zhang2023ochid}
\bibfield{author}{\bibinfo{person}{Shujie Zhang}, \bibinfo{person}{Tianyue Zheng}, \bibinfo{person}{Zhe Chen}, \bibinfo{person}{Jingzhi Hu}, \bibinfo{person}{Abdelwahed Khamis}, \bibinfo{person}{Jiajun Liu}, {and} \bibinfo{person}{Jun Luo}.} \bibinfo{year}{2023}\natexlab{b}.
\newblock \showarticletitle{{OCHID-Fi: Occlusion-Robust Hand Pose Estimation in 3D via RF-Vision}}. In \bibinfo{booktitle}{\emph{Proc. of IEEE/CVF ICCV}}. \bibinfo{pages}{15112--15121}.
\newblock


\bibitem[Zhang et~al\mbox{.}(2022a)]%
        {zhang2022can}
\bibfield{author}{\bibinfo{person}{Shujie Zhang}, \bibinfo{person}{Tianyue Zheng}, \bibinfo{person}{Zhe Chen}, {and} \bibinfo{person}{Jun Luo}.} \bibinfo{year}{2022}\natexlab{a}.
\newblock \showarticletitle{{Can We Obtain Fine-grained Heartbeat Waveform via Contact-free RF-sensing?}}. In \bibinfo{booktitle}{\emph{Proc. of IEEE INFOCOM}}. IEEE, \bibinfo{pages}{1759--1768}.
\newblock


\bibitem[Zhang et~al\mbox{.}(2022b)]%
        {zhang2022quantifying}
\bibfield{author}{\bibinfo{person}{Shujie Zhang}, \bibinfo{person}{Tianyue Zheng}, \bibinfo{person}{Hongbo Wang}, \bibinfo{person}{Zhe Chen}, {and} \bibinfo{person}{Jun Luo}.} \bibinfo{year}{2022}\natexlab{b}.
\newblock \showarticletitle{{Quantifying the Physical Separability of RF-based Multi-Person Respiration Monitoring via SINR}}. In \bibinfo{booktitle}{\emph{Proc. of the 20th ACM SenSys}}. \bibinfo{pages}{47--60}.
\newblock


\bibitem[Zhao et~al\mbox{.}(2021)]%
        {zhao2021point}
\bibfield{author}{\bibinfo{person}{Hengshuang Zhao}, \bibinfo{person}{Li Jiang}, \bibinfo{person}{Jiaya Jia}, \bibinfo{person}{Philip~HS Torr}, {and} \bibinfo{person}{Vladlen Koltun}.} \bibinfo{year}{2021}\natexlab{}.
\newblock \showarticletitle{{Point Transformer}}. In \bibinfo{booktitle}{\emph{Proc. of IEEE/CVF ICCV}}. \bibinfo{pages}{16259--16268}.
\newblock


\bibitem[Zheng et~al\mbox{.}(2020)]%
        {zheng2020v2ifi}
\bibfield{author}{\bibinfo{person}{Tianyue Zheng}, \bibinfo{person}{Zhe Chen}, \bibinfo{person}{Chao Cai}, \bibinfo{person}{Jun Luo}, {and} \bibinfo{person}{Xu Zhang}.} \bibinfo{year}{2020}\natexlab{}.
\newblock \showarticletitle{{V2iFi: In-Vehicle Vital Sign Monitoring via Compact RF Sensing}}.
\newblock \bibinfo{journal}{\emph{Proc. of ACM UbiComp}} \bibinfo{volume}{4}, \bibinfo{number}{2} (\bibinfo{year}{2020}), \bibinfo{pages}{1--27}.
\newblock


\bibitem[Zheng et~al\mbox{.}(2021a)]%
        {zheng2021siwa}
\bibfield{author}{\bibinfo{person}{Tianyue Zheng}, \bibinfo{person}{Zhe Chen}, \bibinfo{person}{Jun Luo}, \bibinfo{person}{Lin Ke}, \bibinfo{person}{Chaoyang Zhao}, {and} \bibinfo{person}{Yaowen Yang}.} \bibinfo{year}{2021}\natexlab{a}.
\newblock \showarticletitle{{SiWa: See into Walls via Deep UWB Radar}}. In \bibinfo{booktitle}{\emph{Proc. of the 27th ACM MobiCom}}. \bibinfo{pages}{323--336}.
\newblock


\bibitem[Zheng et~al\mbox{.}(2021b)]%
        {zheng2021more}
\bibfield{author}{\bibinfo{person}{Tianyue Zheng}, \bibinfo{person}{Zhe Chen}, \bibinfo{person}{Shujie Zhang}, \bibinfo{person}{Chao Cai}, {and} \bibinfo{person}{Jun Luo}.} \bibinfo{year}{2021}\natexlab{b}.
\newblock \showarticletitle{{MoRe-Fi: Motion-robust and Fine-grained Respiration Monitoring via Deep-Learning UWB Radar}}. In \bibinfo{booktitle}{\emph{Proc. of the 19th ACM SenSys}}. \bibinfo{pages}{111--124}.
\newblock


\bibitem[Zheng et~al\mbox{.}(2023)]%
        {zheng2023autofed}
\bibfield{author}{\bibinfo{person}{Tianyue Zheng}, \bibinfo{person}{Ang Li}, \bibinfo{person}{Zhe Chen}, \bibinfo{person}{Hongbo Wang}, {and} \bibinfo{person}{Jun Luo}.} \bibinfo{year}{2023}\natexlab{}.
\newblock \showarticletitle{{AutoFed: Heterogeneity-Aware Federated Multimodal Learning for Robust Autonomous Driving}}. In \bibinfo{booktitle}{\emph{Proc. of the 29th ACM MobiCom}}. \bibinfo{pages}{1--15}.
\newblock


\bibitem[Zheng et~al\mbox{.}(2019)]%
        {widar}
\bibfield{author}{\bibinfo{person}{Yue Zheng}, \bibinfo{person}{Yi Zhang}, \bibinfo{person}{Kun Qian}, \bibinfo{person}{Guidong Zhang}, \bibinfo{person}{Yunhao Liu}, \bibinfo{person}{Chenshu Wu}, {and} \bibinfo{person}{Zheng Yang}.} \bibinfo{year}{2019}\natexlab{}.
\newblock \showarticletitle{{Zero-Effort Cross-Domain Gesture Recognition with Wi-Fi}}. In \bibinfo{booktitle}{\emph{Proc. of the 17th ACM MobiSys}}. \bibinfo{pages}{313--–325}.
\newblock


\end{thebibliography}

\end{document}